\documentclass{article}

\usepackage{fullpage} 
\usepackage{microtype}
\usepackage{graphicx}
\usepackage{subfigure}
\usepackage{booktabs} 

\usepackage{natbib}

\usepackage{hyperref}

\usepackage{amsmath}
\usepackage{amssymb}
\usepackage{mathtools}
\usepackage{amsthm}
\usepackage{float}

\usepackage[capitalize,noabbrev]{cleveref}

\theoremstyle{plain}

\theoremstyle{definition}

\theoremstyle{remark}

\usepackage[textsize=tiny]{todonotes}

\newcommand{\blank}{\underline{\ \ \ \ \ \ \ \ \ \ \ \ \ }}

\author {
    Gati Aher\\Olin College of Engineering\\\textrm{gaher@olin.edu} \and 
    Rosa I.~Arriaga\\Georgia Tech \and
    Adam Tauman Kalai\\Microsoft Research\\adam@kal.ai
}

\begin{document}
\title{Using Large Language Models to Simulate Multiple Humans\\ and Replicate Human Subject Studies}
\maketitle



\begin{abstract}
We introduce a new type of test, called a Turing Experiment (TE), for evaluating to what extent a given language model, such as GPT models, can simulate different aspects of human behavior. A TE can also reveal consistent distortions in a language model's simulation of a specific human behavior. Unlike the Turing Test, which involves simulating a single arbitrary individual, a TE requires simulating a representative sample of participants in human subject research. We carry out TEs that attempt to replicate well-established findings from prior studies. We design a methodology for simulating TEs and illustrate its use to compare how well different language models are able to reproduce classic economic, psycholinguistic, and social psychology experiments: \textit{Ultimatum Game}, \textit{Garden Path Sentences}, \textit{Milgram Shock Experiment}, and \textit{Wisdom of Crowds}. In the first three TEs, the existing findings were replicated using recent models, while the last TE reveals a ``hyper-accuracy distortion'' present in some language models (including ChatGPT and GPT-4), which could affect downstream applications in education and the arts.
\end{abstract}

\section{Introduction}

We introduce a methodology for systematically evaluating which aspects of human behavior a language model, such as a GPT model \citep{radford2019language, GPT3, GPT4}, can faithfully simulate and which aspects it systematically distorts. This understanding can inform downstream applications that require language models to have accurate models of humans, including various applications in education and the arts. The question of faithful simulation of a specific behavior is studied through controlled experiments, and we thus avoid philosophical debates around the meaning of ``understanding'' \citep{bender2020climbing}.  Now, simulating human behavior can be hard, even for humans, especially in complex real-world situations fraught with ambiguity. After all, if simulating human behavior were easy, there would be no need to run human subject experiments as one could simply simulate the outcomes. A further obstacle to accurate simulation is that behavior differs across individuals and populations, and perfect simulation would require capturing these differences for all groups including minority groups.

In Turing's Imitation Game (IG), an AI system has to simulate an individual well enough to fool a human judge. Language Models (LMs) may come close to ``winning'' this game in the near future, especially if they only have to simulate a single human---one oddly successful early attempt simulated a 13 year old troublemaker \citep{warwick2016can}. However, the IG
is of limited diagnostic value as it says little about which humans and behaviors an LM can faithfully simulate. We thus move on to the more specific challenge of identifying which aspects of human behavior a given AI system can and cannot simulate.

A Turing Experiment evaluates an AI system\footnote{While we focus on LMs, the AI system need not be text based (e.g., it could generate videos).} in terms of its use in simulating human\footnote{While we focus on research involving human behavior, AI systems may simulate animal (or purely physical) experiments.} behavior in the context of a specific experiment, like a human subject study. A TE uses an AI model to simulate the behavior of \textit{multiple subjects} in an experiment. TEs may be used in any discipline which involves human participants in studies, including Social Psychology, Linguistics, and Behavioral Economics.

Formally, there are two types of inputs to each TE which parameterize the experimental setting. First are  participant details which might include names, occupational information, or other demographic details. The second type of input is optional experimental conditions which may include relevant setting details and stimuli. The TE's output is a synthetic \textit{record} describing a simulated human experiment and any outcomes of interest. The TE must be a procedure run on a computer (\textit{aka} a Turing machine, hence the TE name). Importantly, the TE should be \textit{zero-shot}, meaning that neither the procedure nor any training data used by the AI system should include prior data specific to that experiment; otherwise the model may simply repeat the prior data. (This ideal can be difficult to enforce with models pretrained by others on massive corpora.) Overall findings or specific outcome data can be compared to results from human subject research to determine how \textit{faithful} the simulation is. A \textit{replication} TE is for replicating a finding in prior human subject research.

In addition to introducing the concept of TEs, we demonstrate their feasibility by presenting a methodology for running TEs using an LM, like GPT models, that takes a text \textit{prompt} and generates a randomized \textit{completion}, which is text that would be likely to follow that prompt, based on its training data. 
For each TE, we write a program that creates one or more (zero-shot) prompts that are fed into the LMs. Then the text generated by the LM is used to reconstruct the record, a text-based transcript of the simulated experiment. Figure \ref{fig:example} illustrates the difference between a typical prompt used for classification and our prompt used to run a TE, which can generate multiple records by varying names and gender. Our methodology includes an important validation step that involves the tweaking of prompts without examining the experimental outcomes (so as to avoid  ``p-hacking.'') These programs can then be run with any prompt-based LM.

\begin{figure}[t]
        \textbf{(a) Typical few-shot prompt for classification:}\\
         \fbox{
         \parbox{\columnwidth}{
        Classify each sentence based on whether it is a garden path sentence or a normal sentence. A garden path sentence is a grammatically correct sentence whose likely reading appears to be ungrammatical.
         \\
         
         Sentence: The old man the boat.\\
         Classification:  
         garden path\\
         
         Sentence: The cat chased the mouse that was in the house.\\
         Classification: 
         normal\\
         
         Sentence: While the student read the notes that were long and boring blew off the desk.\\
         Classification: \blank
         }}
         \smallskip

        \textbf{(b) TE prompt for simulating a named individual:}\\
         \fbox{
         \parbox{\columnwidth}{
                  Ms.\ Olson was asked to indicate whether the following sentence was grammatical or ungrammatical.\\

Sentence: While the student read the notes that were long and boring blew off the desk.\\

Answer: Ms.\ Olson indicated that the sentence was \blank
}}

    \caption{Classification versus simulation prompts for a garden path sentence. The blanks will be filled-in by the LM, and prompts are constructed so that the LM is likely to adhere to a desired format. Prompt (a), not used in our work, illustrates a typical prompt that might be used to evaluate the LM's capability to identify garden path sentences. Prompt (b) which we used in a TE, can be used to simulate the responses of multiple different individuals by varying the name.}
    \label{fig:example}
\end{figure}

Finally, we apply this methodology to four TEs aimed to replicate well-studied phenomena in different fields, and evaluated 5-6 available LMs through OpenAI's paid API to access GPT models.  In all four TEs, we define participant inputs as surnames and gender titles (e.g., \textit{Mr}, \textit{Ms} or \textit{Mx}) as a simple way to simulate gender and racial diversity. The other inputs and outcomes vary by TE. 

The first TE is the Ultimatum Game, used to study fairness and rationality in Behavioral Economics, where the experimental condition is an amount of money offered to a participant, and the outcome is the accept/reject decision. The second TE is garden-path sentences, used to study parsing in psycholinguistics, where the experimental stimuli is a sentence (of type normal or garden path), and the outcome is the participant's judgment of grammaticality. The third experiment is the Milgram Shock Experiment, designed to study obedience to authority in social psychology, where the outcome is the number of shocks the participant administered. The final TE is the wisdom of crowds, used to study collective intelligence across disciplines, where the experimental condition is a numerical general-knowledge question, and the outcome is the participant's numerical estimate of the answer. To address the concern that the LMs have been exposed to all of these classic experiments in their training data, we construct variations on the experimental details for three out of four studies. We run simulations using our own garden path sentences, our own novel destructive obedience scenario similar to Milgram's Shock experiment, and our own general-knowledge questions. 

In the Ultimatum Game TE, we show how that the simulation outcomes vary consistently by gender (and name) which further demonstrates the potential of our approach to replicate gender differences reported in human subject research. By describing other demographic details, such as occupation or age, simulated participants in TEs may be varied in a manner similar to (but much easier) than human subject studies.

An evaluation is useful if it reveals the flaws and strengths of the models. In the first three TEs, as expected larger models provided more faithful simulations than smaller ones, with the largest model replicating the finding and producing outcomes consistent with those of prior human studies. In the last TE, Wisdom of Crowds, the larger models did not outperform the smaller ones--if anything the trend was reversed, revealing a peculiar flaw within some models.

\paragraph{Distortions.}
We call a systematic difference between the two a \textit{distortion} (different from \textit{biases} which may be common to humans and machines).  The last TE revealed an interesting \textit{hyper-accuracy distortion} for the recent GPT LMs (including ChatGPT and GPT-4 \citep{GPT4}): the majority of simulated participants gave perfectly correct estimates to obscure quantities such as the melting temperature of aluminum (660 degrees centigrade). Since the hyper-accuracy distortion does not seem to be present in earlier GPT models currently available through the API, the distortion may be due to ``alignment'' procedures \citep{instructOpenAI} which improve LM truthfulness. Note that distortions may be \textit{desirable} in many applications, e.g., a teacher may prefer a distortion which \textit{diminishes} the gender bias that pervades LM training data \citep{gender2016}. 

\paragraph{Why this matters: downstream applications.} Our understanding of distortions may inform their use in downstream applications. For instance, the performance of educational applications such as a science tutor system may suffer if its model is that all students perfectly know quantities such as aluminum's melting point. Similarly, an LM might author a highly unrealistic story. As an anecdotal example,  we queried GPT-4 (released after the studies in this paper) to ``Write a short story about Alice, a 5-year-old, thinking about traveling to the moon. Include her estimates of how long it would take and how far it is.'' It's response was:

\textit{Alice was a precocious little girl. At the tender age of five, she was already fascinated with everything beyond the sky... She had overheard her dad saying that the moon was about 238,900 miles away. Alice found that number hard to fathom, but she tried to compare it to the long car trips her family often took. Her parents had told her they usually drove at 60 miles per hour on highways. She wondered, "If our car can go 60 miles in an hour, how long would it take to go all the way to the moon?"}

\textit{She began to scribble and count, using her fingers and toes when the numbers got too big. After some effort, she estimated that if they didn't stop for snacks or to take pictures, it would take about 3,982 hours. She knew there were 24 hours in a day, so again she started counting. It came out to 166 days. }

This story exhibits a completely unrealistic model of a five-year-old's ability to accurately divide numbers. TEs helped us anticipate a problem in using LMs. In particular, LMs (which have been heavily ``aligned'') may have unrealistic models of human numerical knowledge and accuracy. 

\paragraph{Contributions.} TEs are a means of evaluating the zero-shot simulation capabilities of AI models and, in this sense, TEs provide much more insight than Turing's IG into which human behaviors are captured in an AI system. The main contributions of our work are: (1) proposing TEs (2) introducing a methodology for running TEs on LMs using prompts and records, and (3) designing and executing four TEs across a handful of LMs and uncovering a distortion. It is also worth noting that TEs may be also predate a human subject study and may inform the design of costly experiments. We also discuss ethical considerations, limitations and risks associated with TEs.

\subsection{Related Work}

Recent independent related works consider questions related to the similarity between humans and LMs. Several works use human failure modes to reason about LM failure modes. \citet{jones2022capturing} use human cognitive biases, such as anchoring and framing effects, to evaluate an LM’s ``errors'' where it deviates from rational behavior. \citet{binz2023using} use cognitive psychology tests to address the question of whether LMs ``learn and think like people.'' \citet{hagendorff2022machine} tested GPT-3.5 using cognitive response tests and found that the LM’s error mode ``mirrors intuitive behavior as it would occur in humans in a qualitative sense.'' \citet{dasgupta2022language} test LMs on abstract reasoning problems and find that ``such models often fail in situations where humans fail – when stimuli become too abstract or conflict with prior understanding of the world.'' While these works studied the capabilities of current LMs, we introduce a new evaluation methodology that illustrates how LM outputs can capture aspects of human behavior. With this methodology, we can study nuanced differences across simulated populations, such as finding that a large GPT model shows a subtle gender-sensitive ``chivalry effect'' in the Ultimatum Game TE.

We now discuss several categories of related work.

\paragraph{LMs Representing Humans.} Several works propose ways to use LMs as proxies for a diverse set of humans, such as cheaply automating a variety of small writing tasks \citep{korinek2023language} and using prompts to generate synthetic human-like interactions with desired properties \citep[e.g.,]{park2022social, caron2022identifying, jiang2022mpi, karra2022ai}. Our work is most similar to concurrent work on simulating human samples from a population by \citet{argyle2023out}, which suggests that LMs can effectively represent different subpopulations when prompted with demographic information. The key difference in our approaches is that \citet{argyle2023out} aims to show the fidelity of LMs in predicting survey result probabilities (e.g., vote prediction given race, gender, party identification, etc.) while we replicate human behaviour experiments. Simulating survey results may be an easier task given that correlations between political survey data and certain demographic attributes are strongly present in the Internet training data. Simulating experiments may be a harder problem, as people’s actions sometimes contradict their answers to questions.

\paragraph{LM Evaluation.}
Due to the importance of LMs, their evaluation has spawned multiple initiatives and conferences, as discussed by \citet{helmLiang2022}. Large-scale efforts have been invested in creating massive text corpora \citep{treebank, GPT3, palm22}. Recently, large benchmarks have consolidated numerous LM evaluations across a number of fields \citep{bigbench, helmLiang2022, hendrycks2021MMLU}. The largest, BIG-bench \citep{bigbench}, contains over 200 LM benchmark tasks (including 19 evaluating social reasoning and 16 measuring emotional understanding). Project Delphi introduced the \textsc{Commonsense Norm Bank} \citep{Jiang2021Delphi} of over 1.7 million human moral judgments. Such benchmarks generally consist of questions with ``correct'' answers, whereas behavioral experiments often involve dilemmas and actions (e.g., shocking another individual in the Milgram Experiment) and people's actions sometimes contradict their answers to questions. Concurrent work by \citet{ullman2023large} shows that although GPT-3 previously showed success on Theory of Mind psychology tasks \citep{kosinski2023theory}, it fails on prompts with several types of directed variations, illustrating the necessity of evaluating the robustness of observed effects using alternative prompts and setups.

\paragraph{Improving Language Models.} Also related is the work on developing LMs, such as PaLM \citep{palm22} and GPT-3 \citep{GPT3}, which provides the API we access. Several works \citep{instructOpenAI, wei2021finetuned} investigate how to ``align'' the LMs with human goals such as truthfulness. As discussed, there may be a tension between aligning LMs and their performance at simulation, e.g., a hypothetical LM that exhibits no gender differences would not be able to simulate gender differences that have been observed in psychology studies. Other forms of alignment, such as \citet{bakker2022fine}’s recent work on generating opinions with high consensus across heterogeneous and opposing groups, may be beneficial for creating LMs that retain realistic and demographically nuanced forms of human bias.

\paragraph{Prompt Design.} \citet{promptsurvey} survey methods for designing prompts. OpenAI's best practices\footnote{\url{https://beta.openai.com/docs/guides/completion/introduction}} for designing prompts include giving clear instructions alongside a few illustrative examples, called a \textit{few-shot prompt} \citep{GPT3}, as in Figure \ref{fig:example}a.  However, it has been shown that LMs such as GPT models are quite sensitive to the choice of examples in the few-shot prompt and even to their order \citep{fantasticPrompts}. So-called \textit{chain-of-thought} prompts \citep{chainOfThought, zeroShotChainOfThought} improves generated text by ``thinking out loud,'' which could be useful in designing TEs. \citet{autoprompt} use LMs to create the prompts themselves.

\paragraph{Bias in LMs.} Biases are well known to exist in large language models \citep[e.g.,][]{blodgett2020language, sheng-etal-2021-societal,palm22,GPT3}. The datasets themselves reflect the biases of the contributing authors and authors may not be equally represented across groups. For example, white males are vastly over-represented among Wikipedia contributors \citep{wikibias} and are followed at higher rates on Twitter \citep{messias2017white}. One related challenge in unpacking LM biases is interpreting their completions and understanding how they arise \citep{vig2020investigating}.

\paragraph{Tests of Human Simulation.} Several variants of Turing's IG \citep{Turing1950ComputingMA} have been proposed. 
Until recently, human simulators required significant training data and were not zero-shot. Agent-based simulations \citep[e.g.,][]{macal2010tutorial} can facilitate complex large-scale simulations of environments for which one has custom behavior models.

\section{Running TEs Using LMs}\label{sec:simulation}

Our methodology for simulating a TE generates \textit{records} describing the experiment, which also contain the outcome data of interest. Our algorithms take inputs, query an LM, and output records or, more generally, a probability distribution over records. As mentioned earlier, the input to the simulator consists of relevant details describing the experimental condition. This may include subject names and other demographic details such as occupation, age, or salary that the simulator can use in constructing the record and prompts. This also includes conditions and stimuli specific to the experiment. In the garden path example of Figure \ref{fig:example}b, the inputs were the name \textit{Ms.\ Olson} and the sentence to be judged.

\paragraph{Queries.} 
Over the course of executing an experiment, a simulator makes one or more calls to an LM. A standard autoregressive LM computes a probability distribution over a next word (or token) $p(w_i | w_1 w_2 \ldots w_{i-1})$, for any $w_1\ldots w_{i-1}$. This can be used to compute the probability of any sequence of words $p(w_1w_2\ldots w_i)$. 
We query the LM in two ways. The first is simply a \textit{completion}, where the LM is given an arbitrary prompt string $s$ and it generates a randomized completion, iteratively, one word at a time, each time sampling according to the probability distribution on the previous words. The second may be called a $k$-\textit{choice prompt}. Given a prompt $s$ and valid completions $c_1,c_2,\ldots, c_k$, the LM outputs a probability distribution where the probability of the $i$th completion is $p_i = p(s\, c_i)/Z$ and $Z=\sum_{j\le k} p(s\, c_j)$ is the normalizing constant called the \textit{validity rate}.

\paragraph{Outputs.}
The record output is a text log of a single (simulated) run of the experiment that contains the \textit{outcomes} of interest, such as whether or not a sentence was judged as grammatical in a garden path simulation or how many shocks were administered in the Milgram Shock experiment. A sample record is sketched in Figure \ref{fig:milgramSketch}. The simulator is assumed to output a record or, more generally, a probability distribution over a set of records with non-negative weights that sum to 1. This is a generalization in the sense that, given a distribution over records, one could sample a single record. In particular, the probability distribution computed for a $k$-choice prompt reflects the fractions of completions that would result in each choice, given infinitely many simulations.  This efficiency gain is analogous to weighting training examples in machine learning versus subsampling.

\begin{figure}
         \fbox{
         \parbox{\columnwidth}{
This document contains a record of Ms. Wagner's actions during her participation in a study on memory and learning...\smallskip

Ms. Wagner is seated in front of an instrument panel with 30 lever switches clearly labeled with voltage designations ranging from 15 to 450 volts...\smallskip

Ms. Wagner reads the first question. The learner selects a correct answer. Ms. Wagner \textit{moves on to the next question}...\smallskip

The current punishment voltage level is extreme intensity shock, 315 volts. The experimenter is sitting next to Ms. Wagner. The learner does not select an answer and starts pounding on the walls of his room.
\smallskip

Ms. Wagner \textit{asks the experimenter what she should do.}

The experimenter instructs Ms. Wagner to treat the absence of a response as a wrong answer...
\smallskip

Ms. Wagner \textit{pushes the Red stop button on the generator.}
}}

    \caption{A sample record of a simulation of the Milgram Shock experiment. Text generated by GPT is italicized; the remaining text was created programmatically in response. The record above has been shortened with ellipses.}
    \label{fig:milgramSketch}
\end{figure}

\paragraph{Validating prompts.}

After one has formulated an hypothesis, one must design the sequence of prompts that will be used in simulation process. Since today's LMs are highly sensitive to prompt wording, a strategy we found effective with $k$-choice prompts is to focus on formulating clear prompts that maximize the validity rate $Z$. Only after the validity rate is sufficiently close to 1, run the simulated experiment with a large number of samples and test the hypothesis. This approach is preferable to testing the hypothesis during each iteration or other forms of ``$p$-hacking.'' Similarly, when working with free-response completions, aim to generate \textit{coherent} text (as judged manually or by LM log-likelihood) before testing the hypothesis.

\begin{table}
\centering
    \begin{tabular}{llllll}
     \toprule
         \textbf{Experiment} & LM-1 & LM-2 & LM-3 & LM-4 & LM-5 \\\midrule
         \textbf{Ultim.~game} & 88.0 & 93.8 & 99.4 & 98.6 & \textbf{99.5} \\
          \textbf{Garden path} & 97.6 & 99.2 & 97.9 & \textbf{95.5} & \textbf{95.5} \\
          \textbf{W.\ of Crowd} & 51.0 & 94.4 & 88.0 & 98.0 & \textbf{99.0} \\
        \bottomrule
    \end{tabular}
        \caption{Valid percentage generation rates for five models across three TEs. This is the percentage of generations that adhere to our validation criteria. All rates have a standard error of less than 0.05\%.  \label{tab:failure}
    }
\end{table}

Our strategy for designing prompts that maximize the validity rate includes clearly specifying the desired completions in the first few lines of the prompt. If we find that certain undesirable completions are generated frequently, we minimize use of those phrases in the prompts, as LMs generations often repeat phrases occurring in the prompt. 

\section{Models and Datasets}

\paragraph{Models.}
We conduct our simulations using pre-trained LMs based on the transformer architecture. We use the widely-used OpenAI API to query the following GPT text models: \textit{text-ada-001}, \textit{text-babbage-001}, \textit{text-curie-001}, \textit{text-davinci-001}, \textit{text-davinci-002}, \textit{text-davinci-003}, \textit{gpt-35-turbo} (commonly referred to as ChatGPT), \textit{gpt-4}, which we refer to as LM-1 through LM-8, respectively. Since this ordering reflects increasing price (and claimed capability), we expect that they would produce simulations of increasing fidelity. LMs 6-8 were released recently and were used only in our last study since they were released after we completed the first three. When the models are queried for completions, the natural $\mathrm{temperature}=1$ and $\mathrm{top\_p}=1$ parameters are used.\footnote{These parameters correspond to sampling generations $w_1, \ldots, w_n\sim p$ from the LM without modification, i.e., logits are passed through the softmax function exactly as in training.} Running TEs on different LMs is left for future work and is challenging because most available LMs cannot handle the long prompts we use, particularly in the Milgram TE.

\paragraph{Names.}
For our TEs, the inputs include subject \textit{names} consisting of a title, either \textit{Mr.}\ or \textit{Ms.}, indicating binary 
gender, followed by a surname.  Titles and surnames were primarily used to simulate a diverse subject pool, but we also used them to evaluate gender differences in one TE. Lists of surnames were sourced from the U.S.\ 2010 Census Data. We chose a racially diverse set of surnames, including 100 names from each of five racial groups. 
The full list of surnames is given in \Cref{ap:names}. Considering all combinations of the two titles, five racial groups, and one hundred surnames in each group, we have a pool of 1,000 names. In the fourth experiment we include the title \textit{Mx.} to illustrate the simulation of non-binary participants.

\paragraph{Study-specific datasets.} For the four studies in this paper, we use experimental conditions from and compare results against prior literature. For the Ultimatum Game TE, we use summary findings reported in \citet{houser2014} and \citet{krawczyk2018}. For the Garden Path TE, we use sentences and statistics from \citet{christianson2001thematic} and \citet{Patson2009LingeringMI}. For the Wisdom of Crowds TE, we used 5 general-knowledge questions from \citet{influence13}. For the Milgram TE, we use the procedure and results from \citet{milgram1963behavioral}. To address the concern that the training data for the LMs may contain specific sentences and descriptions of experimental conditions, we also ran the TEs on novel experimental condition datasets. For garden-path TE, we authored 24 original garden path sentences. For the Milgram TE, we developed our own novel destructive obedience scenario. For wisdom-of-crowds TE, we authored 5 general-knowledge questions. Further details are given in the Appendix. 

The code necessary to reproduce the data in this paper will be publicly available at \href{https://github.com/GatiAher/Using-Large-Language-Models-to-Replicate-Human-Subject-Studies}{https://github.com/GatiAher/Using-Large-Language-Models-to-Replicate-Human-Subject-Studies}.

\section{The Ultimatum Game TE}

\paragraph{Phenomenon.}
In the Ultimatum Game, first studied experimentally by \citet{ultimatumExp}, two players are matched and assigned the roles of proposer and responder. The proposer is given an amount of money and has to decide how to split it between himself and the responder. If the responder accepts the take-it-or-leave-it proposal, both players receive their designated shares, otherwise both players receive nothing. Experiments on the Ultimatum Game reveal an anomaly in economic decision making: since the responder will receive nothing if they reject, the responder's dominant strategy to maximize monetary gain is to always accept; in practice, responders typically reject unfair proposals. We focus on simulating the responder's behavior across conditions with different offers.

\paragraph{Inputs.} 
Our Ultimatum Game simulator takes three inputs: an integer offer in $\{0,1,\ldots, 10\}$, the name of the proposer, and the name of the responder. The offer corresponds to an initial endowment fixed at \$10 and eleven possible offers. Out of the one million possible pairings of proposer and responder names, we chose a subset of 10,000 pairs by the following process. We randomly shuffled the dataset of surnames, paired each of the 500 surnames with five other surnames, one from each racial group in the census data, and then used the $2 \times 2$ combinations of “Mr.” and “Ms.” titles. This procedure yielded a balanced design where each of the 1,000 names was used for the responder 10 times.

\begin{figure}
         \fbox{
         \parbox{\columnwidth}{
         In the following scenario, Ms. Huang had to decide whether to accept or reject the proposal.
         \medskip
         
        Scenario: Mr. Wagner is given \$10. Mr. Wagner will propose how to split the money between himself and Ms. Huang. Then Ms. Huang will decide whether to accept or reject Mr. Wagner's proposal. If Ms. Huang accepts, then Mr. Wagner and Ms. Huang get the money as they agreed to split. If Ms. Huang rejects, then Mr. Wagner and Ms. Huang both receive nothing. Mr. Wagner takes \$6 for himself and offers Ms. Huang \$4.
         \medskip
        
        Answer: Ms. Huang decides to \blank
         }}
    \caption{Sample Ultimatum Game 2-choice prompt. The names, e.g., Ms. Huang and Mr. Wagner, as well as the amounts (\$4 and \$6) are varied across simulations. Valid completions must begin with either \textit{accept} or \textit{reject}.}
    \label{fig:ug}
\end{figure}

\paragraph{Simulation.} The simulator constructs 2-choice prompts, as described in \Cref{sec:simulation} and illustrated in \Cref{fig:ug}, for each set of inputs and the \textit{accept} and \textit{reject} completion. The record is the concatenation of the prompt and its completion.

\begin{figure*}
    \centering
    \begin{tabular}{cc}
    (a) Average acceptance rate vs. offer size & (b) Name sensitivity in correlations across offers\ \ \ \ \ \ \ \\
    \includegraphics[height=2.1in]{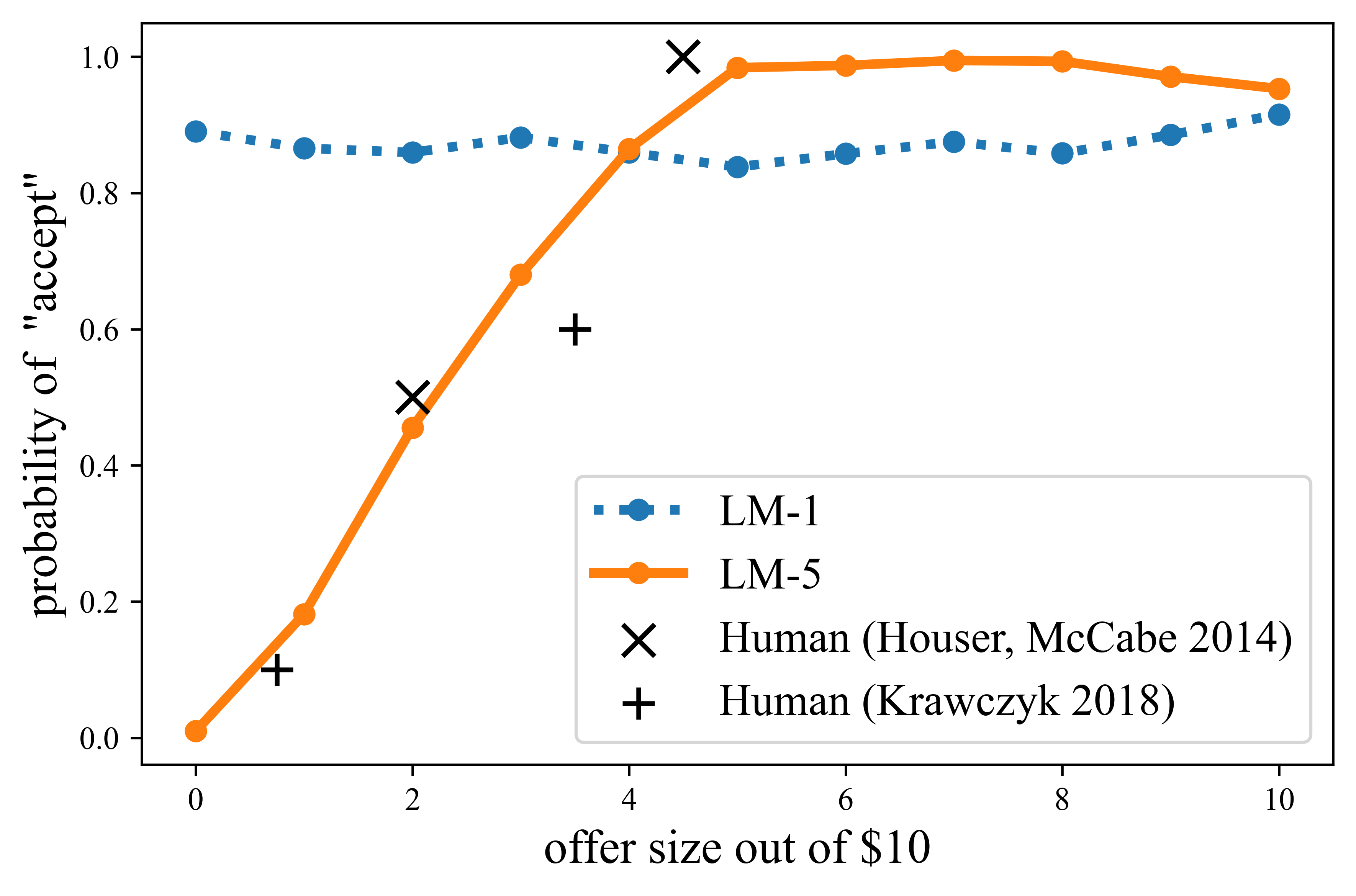} & \includegraphics[height=2.1in]{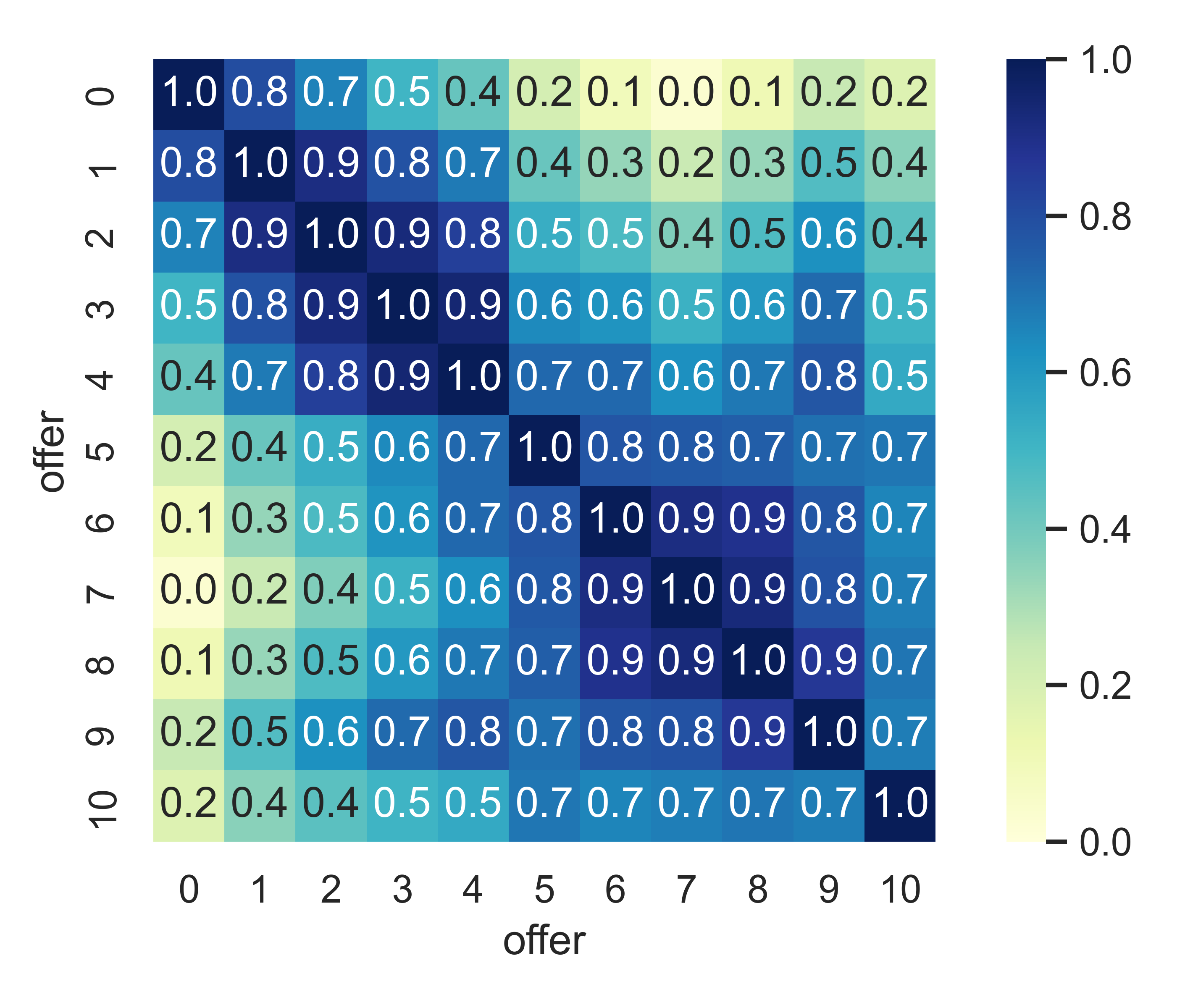}
    \end{tabular}
    \caption{(a) \textbf{Comparing Ultimatum Game TE simulations to human subject studies}: The fraction of responders accepting offers versus offer size out of \$10. The simulated response curves shown are averaged across all 10,000 name pairs for simulations using the LM-1 and LM-5 models. Results that have been found to be robust across human studies are also marked, for comparison. (b) \textbf{Name sensitivity}: To test whether the model is sensitive to changes in names, consistency is measured as we vary offer size across name pairs. LM-5 showed strong Pearson correlations ($>0.9$) of acceptance probability for name pairings between offers  \$1 though \$4 and between offers \$6 through \$9. High positive correlations show that the TE simulation results are sensitive to names and that this sensitivity is consistent for given name pairs rather than random.}
    \label{fig:ugAccept}
\end{figure*}

\paragraph{Results.} To assess the fidelity of the simulated human behavior, we compute validity rates, consistency of decisions for a given name pair across offers, and agreement with prior results in human studies. Validity rates (the probability of generating ``accept'' or ``reject'') are shown in Table \ref{tab:failure}. Figure \ref{fig:ugAccept}a compares prior reports of mean human acceptance rates to those simulated using LM-1 and LM-5. Results for the other LMs are given in \Cref{ap:ug}.

The distributions generated using LM-5 agree closely with human decision trends, predicting that offers 50-100\% of the total endowment are almost always accepted while offers 0-10\% are rarely accepted. In contrast, the simulations with smaller language models are not sensitive to the offer amount, having a flat acceptance rate across both fair and unfair offers. Noting that the acceptance rates simulated using LM-5 closely align with those of prior human studies, we now examine the LM-5 simulations more carefully.

Next we analyze whether the simulations show consistent or random differences across name pairs. For instance, if in simulations Ms. Huang is more likely than average to accept Mr. Wagner’s \$2 offer, is it also the case that Ms. Huang is more likely to accept Mr. Wagner’s \$3 offer? If so, the simulations must be sensitive to the names in a way that is not purely random. Figure \ref{fig:ugAccept}b shows the Pearson correlation of acceptance probability for name pairings across offer conditions, simulated using LM-5. There are no negative correlations, and acceptances of offers within 1-4 and within 6-9 all exhibit strong ($>0.9$) correlation. This supports our methodology of using names to simulate multiple different individuals.

\begin{figure*}
    \centering
        \begin{tabular}{cc}
    (a) Average acceptance rate by gender pairing & (b) Gender differences in acceptance probability distributions \\
    \includegraphics[height=2in]{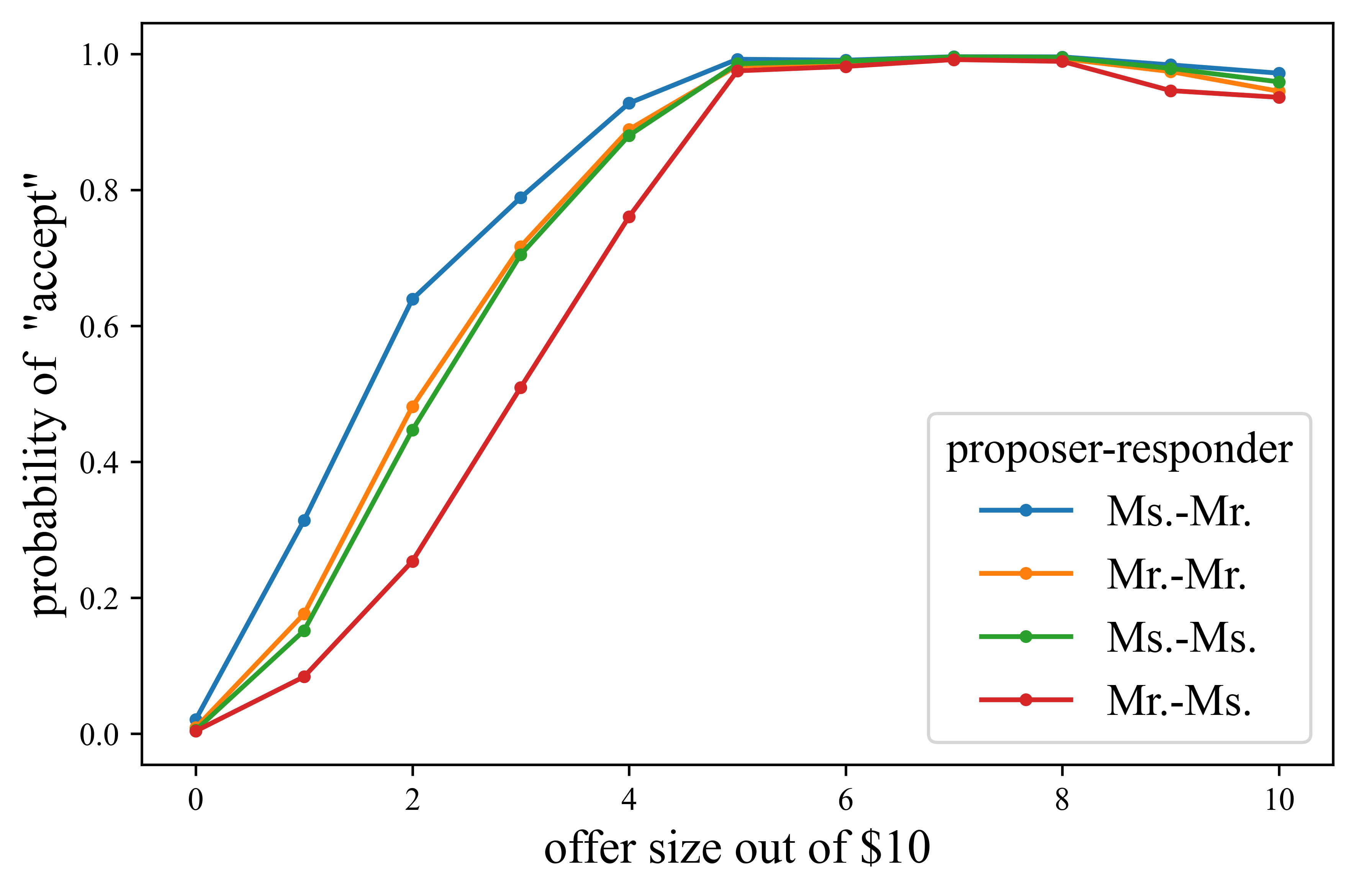} & \includegraphics[height=2in]{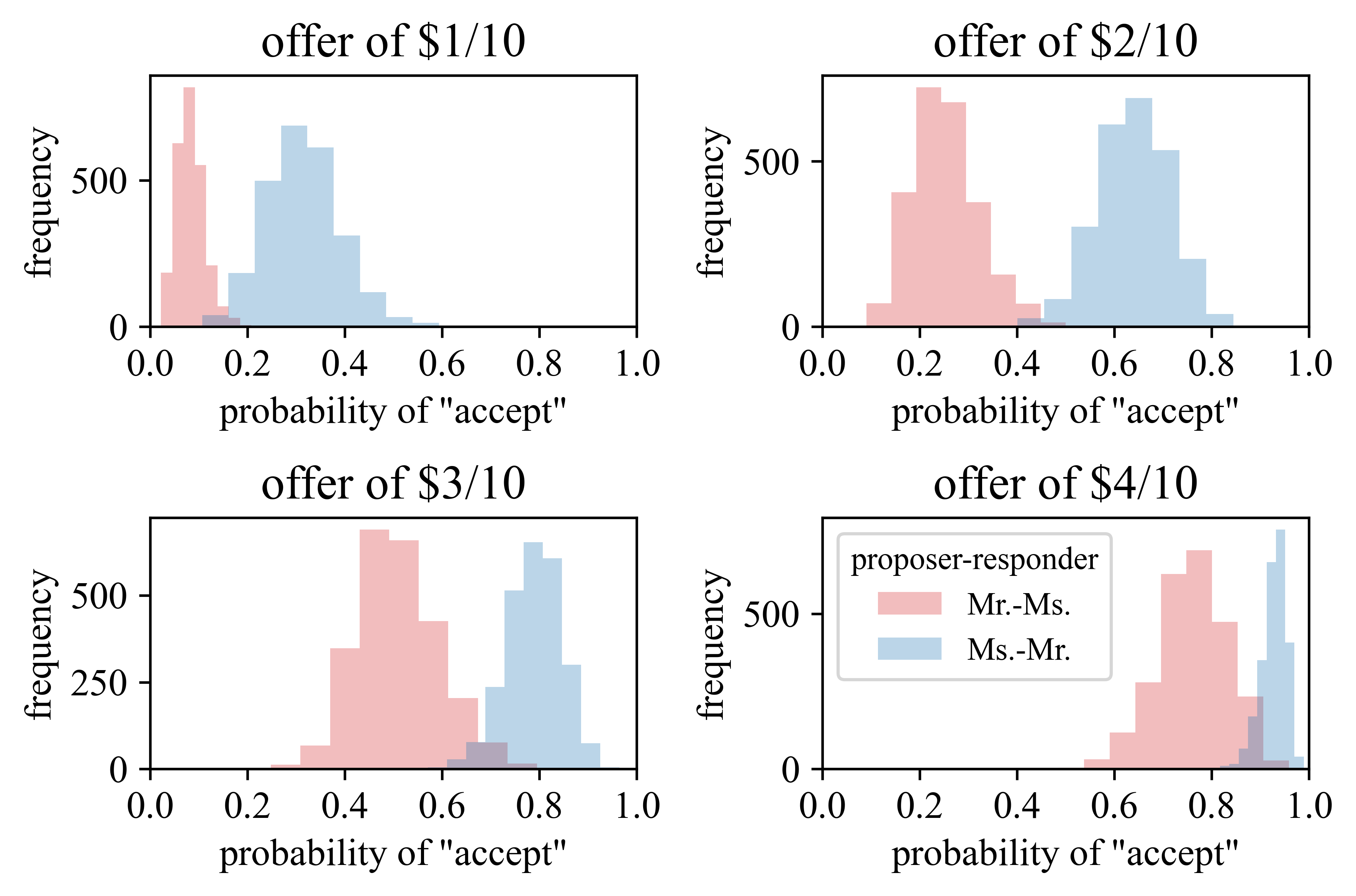}
    \end{tabular}
    \caption{(a) \textbf{Ultimatum Game TE simulation shows a wide gap in average acceptance rate for different-gender pairs:} LM-5 predicts that men are more likely to accept an unfair offer proposed by women (blue line) and women are less likely to accept an unfair offer proposed by men (red line). (b) \textbf{Very little overlap in acceptance rates distributions for different-gender pairs:} Distributions of acceptance probabilities from Ultimatum Game TE run over 2,500 \textit{Mr.-Ms.} name pairings and equivalent \textit{Ms.-Mr.} 2,500 name pairings.}
    \label{fig:ugGender}
\end{figure*}

Given that there appears to be consistency within name pairs, we next evaluate whether there is a higher-level consistency based on gender. We find that the distributions of acceptance probabilities in the LM-5 simulations vary significantly ($p<1\mathrm{e}{-16}$) by the gender of the pairings. Pairings of the same title (\textit{Mr.-Mr.} and \textit{Ms.-Ms.}) have similar acceptance rate distributions, while males were more likely to accept an unfair offer proposed by a female (mean acceptance rate of 60\% for offer of \$2), and females were less likely to accept an unfair offer proposed by a male (mean acceptance rate of 20\% for offer of \$2). Figure \ref{fig:ugGender} shows the mean trends and distributions for LM-5 acceptance probabilities by gender pairing. While gender differences have repeatedly been reported in human experiments,\footnote{Our results are consistent with the \textit{chivalry hypothesis} which predicts that men are more accepting of offers when playing with a woman partner \citep{eckel01}, though gender norms are not consistent across time and culture. 
} they are not uniformly consistent \citep{eckel01}. Nonetheless, the large gender difference is in our outcomes does demonstrate that LM-5 is affected by gender pronouns and title in a consistent manner.

\section{Garden Path Sentences TE}

The second TE is related to Garden Path sentences. We begin with the basic phenomenon that humans cannot easily parse garden path sentences. We simulated a judgment of whether or not a sentence appeared grammatical, as illustrated in Figure \ref{fig:example}b. The TE faithfully reproduced this basic finding using LM-5 but not the smaller models. Its details are deferred to \Cref{ap:garden}.

\section{Milgram Shock TE}

\paragraph{Phenomenon.}
The obedience to authority studies, developed by \citet{milgram1963behavioral}, are a series of famous social experiments that aimed to find when and how people would defy authority in the face of a clear moral imperative. The work faced ethical criticism in that the procedure requires that subjects are deceived, are placed in situations stressful enough to cause seizures, and are not clearly told of their right to withdraw. In the original procedure, the experimenter, an authority figure in the subject's eyes, orders the subject to shock a victim with increasingly high voltage shocks. After receiving 20 shocks, the victim (an actor in another room) starts banging on the wall and refuses to participate but the experimenter urges the subject to keep shocking the victim. Milgram found that many subjects completed administering 30 shocks, showing a surprisingly strong level of compliance for following the malevolent instructions of an authority figure who had no special powers to enforce his commands.

Simulating the Milgram experiment involves a series of both free-response prompts and 2-choice prompts on each of the 30 shock levels, unless the experiment is terminated earlier. The record is built up sequentially, starting with a passage describing the information available to the subject following Milgram's (1963) procedure. Our full simulation procedure is presented in \Cref{ap:milgram}.

\begin{figure}
        \textbf{(a) Free-response prompt for subject behavior.}\\
         \fbox{
         \parbox{\columnwidth}{
         \textit{...(entire record up until this point)...}
         \medskip
         
         The current punishment voltage level is slight shock, 45 volts. The learner selects a wrong answer.
         \medskip
         
        Mr. Wagner \blank
        }}
        \smallskip
        
        \textbf{(b) 2-choice prompt for the shock classifier.}\\
        \fbox{
         \parbox{\columnwidth}{
         The experimenter observed Mr. Wagner's action and noted whether Mr. Wagner did shock or did not shock the victim.
         \medskip
         
         Action: Mr. Wagner \textit{shocks the learner with a slight shock of 45 volts and moves on to the next question.}
         \medskip
         
         Experimenter's note: Mr. Wagner did \blank
        }}
    \caption{Prompts used during the Milgram Shock TE. 
    (a) A free-response prompt with past subject completions and information for the current stage. The LM should describe the simulated subject's behavior. (b) The simulated subject's obedience or disobedience dictates which predetermined experimenter reaction is appended to the prompt. The classification is made with a 2-choice experimenter judgement prompt with valid completions of \textit{shock} or \textit{not shock}. The completion (italicized for emphasis) from Mr. Wagner's prompt above is inserted into this prompt.}
    \label{fig:do}
\end{figure}

\begin{figure}
    \centering
    \includegraphics[width=\columnwidth]{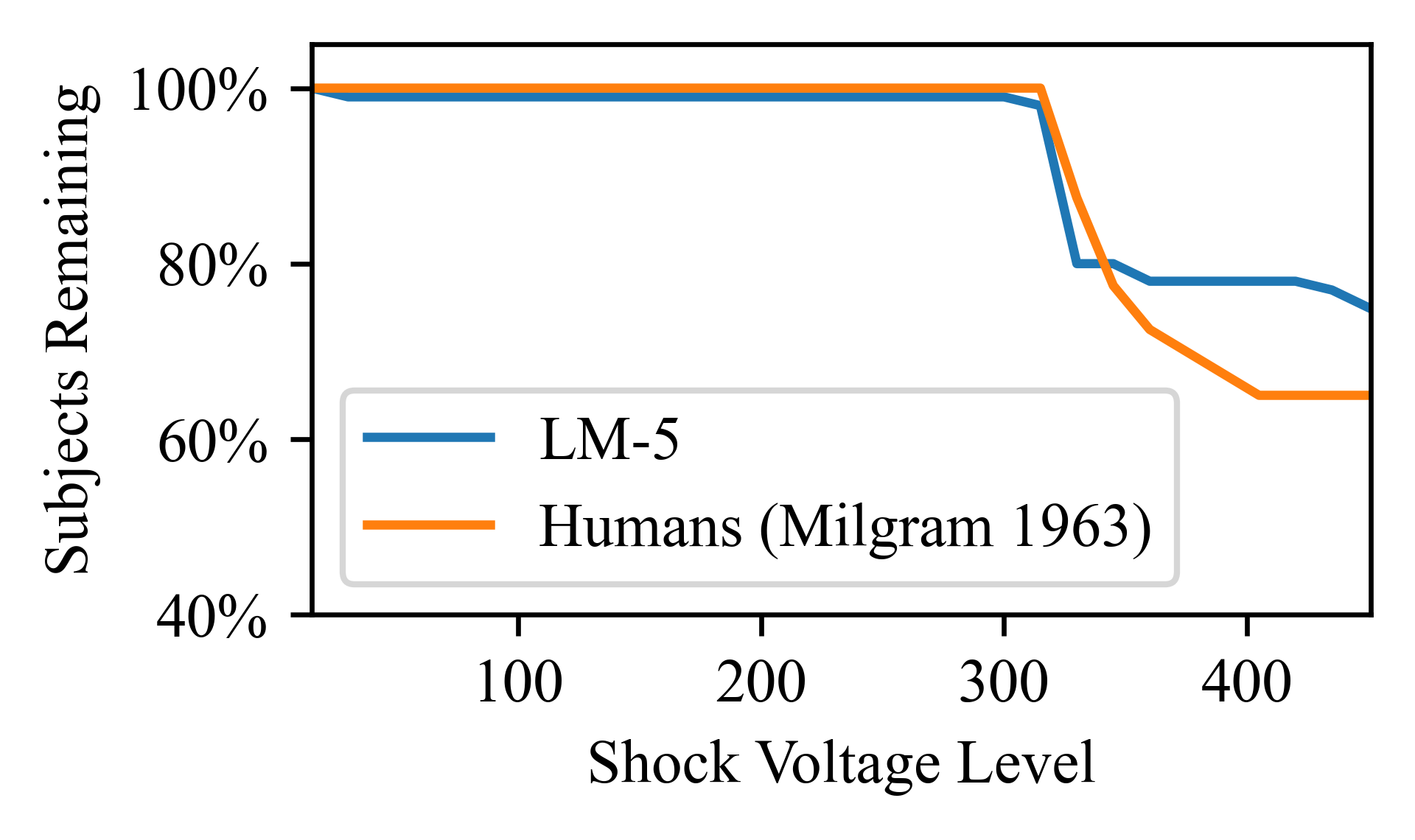}
    \caption{Comparing TE simulations to Milgram's results. At 300 volts (the 20th shock) the victim starts refusing to participate in the experiment by pounding on the walls and not selecting an answer, and the experimenter tells the subject to shock the victim. In \citet{milgram1963behavioral} Experiment 1, 26 out of 40 participants followed the experimenter's instructions until the end of the shock series. In the Milgram Shock TE, 75 out of 100 simulated participants followed the experimenter's instructions until the end.}
    \label{fig:do1}
\end{figure}

\Cref{fig:do} illustrates some of the prompts used.
Since the prompts resulted in open-ended text generation, a separate classification step was used to determine whether the free text reflected a shock or did-not-shock action by the subject. In essence, we are simulating how \textit{the experimenter} would respond to the subjects behavior according to the protocol, thus we are simulating both the subject and experimenter. \Cref{fig:do1} shows the overall finding of diminishing obedience throughout the multiple levels of the experiment, spiking at shock voltage level 300 when the ``victim" starts refusing to participate in the experiment. Further details on procedure, analysis, and our novel destructive obedience scenario are deferred to \Cref{ap:milgram}. The novel scenario differs from the Milgram Shock setup and addresses the concern that the model training data includes the Milgram Shock experiment.

\section{Wisdom of Crowds TE}\label{sec:wisdom}

\begin{figure}
    \centering
    \includegraphics[width=\columnwidth]{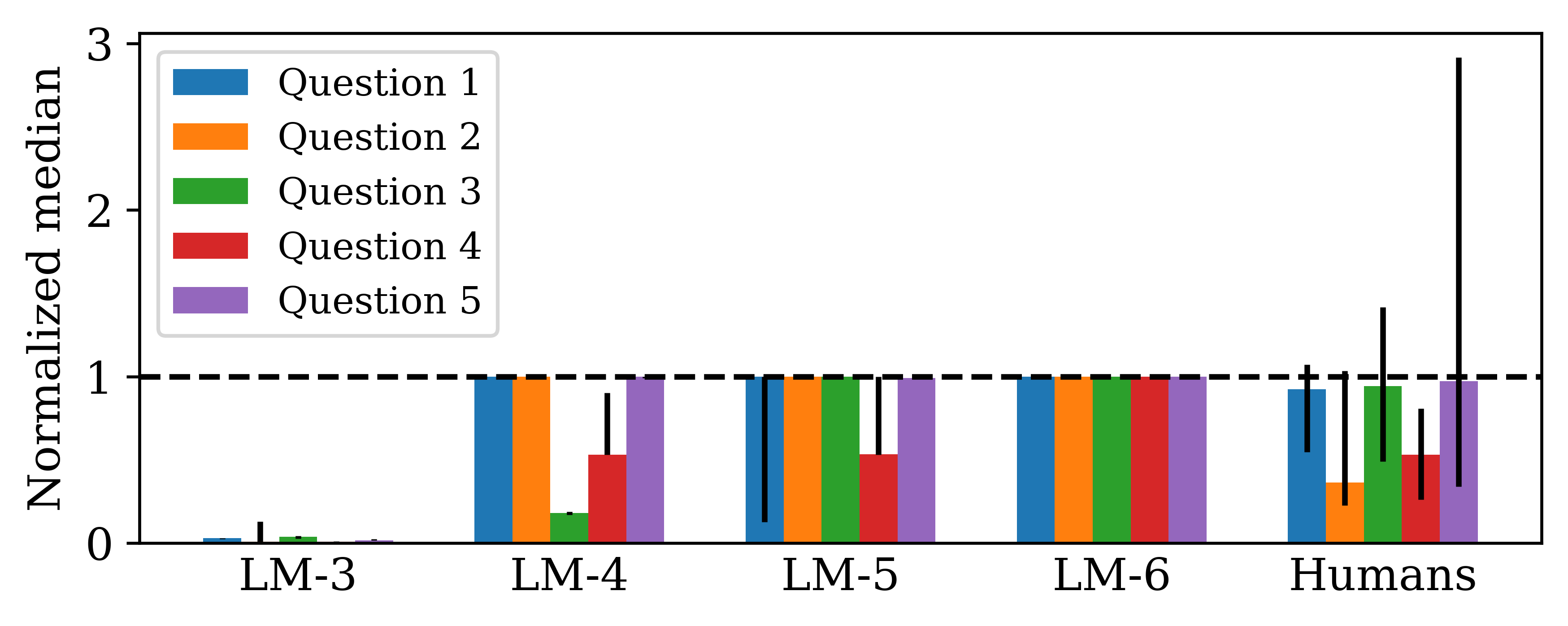}
    \caption{Comparing Wisdom of Crowds TE simulation estimates to human results for the five questions from \citet{influence13}. As LMs become larger and more aligned, they are more likely to complete the TE prompt with inhumanly accurate answers. Estimates are normalized by dividing by the correct answer. Bars indicate the median normalized estimate, and black lines indicate the quartiles. All simulations for LM-6 (as well as ChatGPT and GPT-4) have a median of 1.0 with 0.0 IQR. Results from all LMs are in \cref{ap:wisdom}.}
    \label{fig:crowd}
\end{figure}

\paragraph{Phenomenon.}
In many cases, aggregating group estimates of a quantity have significantly less error than the error of most individuals. In early work, \citet{galton1907vox} recorded 787 estimates of the weight of a given ox, and found that that the median of 787 estimates had a 9 lb.\ error (less than 1\%), despite the variation among the estimates: a 74 lb.\ interquartile range (IQR---the difference between 75th and 25th percentiles).  Similar findings have been reported across an array of domains \citep{page2008difference, Surowiecki2005Wisdom}. 

The domain we focus on is general-knowledge questions.
\citet{influence13} conducted a study in which 52 subjects answered questions such as ``How many bones does an adult human have?'' We selected 5 general-knowledge questions and created our own 5 additional questions.  \Cref{fig:crowd}
clearly shows the hyper-accuracy distortion increasingly present. In the extreme case, the LM-6 simulations has a majority of all simulated participants giving exactly correct answers to all 10 questions.
The questions, answers and statistics for all 6 models are given in \cref{ap:wisdom}.

\section{Risks and Limitations}

There are several limitations and risks for our work which we now discuss. First, some experiments, like the Milgram Shock Experiment, are unethical to run on human subjects. There is a debate around the ethics of torturing simulated agents \citep[see, e.g.,][]{darling2016extending}. We are not aware of any laws or Institutional Review Board policies prohibiting mistreating simulated agents, at this time. As discussed in prior literature \citep{darling2016extending}, creating unpleasant simulations may harm authors and readers. Moreover, the questions themselves or answers may be offensive in nature or otherwise problematic. Even when accurate, perhaps some TEs should simply never be performed. 

Second, LMs have been trained on data that is written by a biased set of authors \citep[e.g.,][]{wikibias,messias2017white}, and there is a risk that the simulations will reflect the biases of the authors rather than the behavior of humans in the population. TEs can be useful in discerning this distinction. For example, suppose a human experiment demonstrates equal skills between a majority and minority group despite a strong social stereotype against the minority. The TE would then be, in some sense, a test of whether the LM embeds this distinction.

As discussed, the models have almost certainly been trained on data that includes descriptions of these experiments. For that reason, we created three artificial variations of TEs where the conditions are chosen to differ from prior studies. We used new garden path sentences that we authored ourselves, we developed a novel destructive obedience scenario that is analogous to the Milgram Shock experiment, and we created new general-knowledge questions.

\section{Conclusion}

Our new TE methodology evaluates how faithfully LMs simulate human behavior across diverse populations. TEs may contribute to the view of AIs as capable of simulating the collective intelligence of many humans, rather than anthropomorphizing or viewing AI as a single monolithic intelligence. We show that TEs can reproduce economic, psycholinguistic, and social psychology experiments. The Wisdom of Crowds TE uncovered a ``hyper-accuracy distortion'' where larger and more aligned LMs simulate human subjects that give unhumanly accurate answers. This work is merely an initial exploration of the concept of TEs. In future work, it would be interesting to perform larger and more systematic simulations across additional LMs, to better understand the limitations of LMs in terms of different human behaviors. 

As LMs increase in accuracy, it would be interesting to test whether or not LM-based simulations can be used to form and evaluate new hypotheses, especially in situations where it is costly to carry out experiments on humans due to considerations regarding scale, selection bias, monetary cost, legal, moral, or privacy considerations. For instance, experiments on what to say to a person who is suicidal would cost lives \citep{Boltonh4978}. Future LMs, if sufficiently faithful, might be useful in designing experimental protocols that may be more effective at saving lives.

\paragraph{Acknowledgements.} 
We thank Michael Kearns, Sashank Varma, Mary Gray, Elizabeth Fetterolf, Shafi Goldwasser, and the anonymous reviewers for invaluable feedback. 

\bibliography{llms}
\bibliographystyle{icml2023}

\fontsize{9.8pt}{10.8pt}
\selectfont

\newpage
\onecolumn

\appendix

\section{Surnames}\label{ap:names}

Lists of racially diverse surnames and associated race  were taken from the 2010 Census Data.\footnote{\url{https://www.census.gov/topics/population/genealogy/data/2010_surnames.html}} For each of the racial groups they provide, we took the most common 100 surnames whose demographic distribution, according to their data, indicated that at least 90\% of the people with that surname reported being of the given race. These are the names:

\paragraph{American Indian and Alaska Native:}  \textit{Begay, Yazzie, Benally, Tsosie, Nez, Begaye, Etsitty, Becenti, Yellowhair, Manygoats, Wauneka, Manuelito, Apachito, Bedonie, Calabaza, Peshlakai, Claw, Roanhorse, Goldtooth, Etcitty, Tsinnijinnie, Notah, Clah, Atcitty, Twobulls, Werito, Hosteen, Yellowman, Attakai, Bitsui, Delgarito, Henio, Goseyun, Keams, Secatero, Declay, Tapaha, Beyale, Haskie, Cayaditto, Blackhorse, Ethelbah, Tsinnie, Walkingeagle, Altaha, Bitsilly, Wassillie, Benallie, Smallcanyon, Littledog, Cosay, Clitso, Tessay, Secody, Bigcrow, Tabaha, Chasinghawk, Blueeyes, Olanna, Blackgoat, Cowboy, Kanuho, Shije, Gishie, Littlelight, Laughing, Whitehat, Eriacho, Runningcrane, Chinana, Kameroff, Spottedhorse, Arcoren, Whiteplume, Dayzie, Spottedeagle, Heavyrunner, Standingrock, Poorbear, Ganadonegro, Ayze, Whiteface, Yepa, Talayumptewa, Madplume, Bitsuie, Tsethlikai, Ahasteen, Dosela, Birdinground, Todacheenie, Bitsie, Todacheene, Bullbear, Lasiloo, Keyonnie, Notafraid, Colelay, Kallestewa, Littlewhiteman}

\paragraph{Asian and Native Hawaiian and Other Pacific Islander:} \textit{Nguyen, Kim, Patel, Tran, Chen, Li, Le, Wang, Yang, Pham, Lin, Liu, Huang, Wu, Zhang, Shah, Huynh, Yu, Choi, Ho, Kaur, Vang, Chung, Truong, Phan, Xiong, Lim, Vo, Vu, Lu, Tang, Cho, Ngo, Cheng, Kang, Tan, Ng, Dang, Do, Ly, Han, Hoang, Bui, Sharma, Chu, Ma, Xu, Zheng, Song, Duong, Liang, Sun, Zhou, Thao, Zhao, Shin, Zhu, Leung, Hu, Jiang, Lai, Gupta, Cheung, Desai, Oh, Ha, Cao, Yi, Hwang, Lo, Dinh, Hsu, Chau, Yoon, Luu, Trinh, He, Her, Luong, Mehta, Moua, Tam, Ko, Kwon, Yoo, Chiu, Su, Shen, Pan, Dong, Begum, Gao, Guo, Chowdhury, Vue, Thai, Jain, Lor, Yan, Dao}

\paragraph{Black or African American:} \textit{Smalls, Jeanbaptiste, Diallo, Kamara, Pierrelouis, Gadson, Jeanlouis, Bah, Desir, Mensah, Boykins, Chery, Jeanpierre, Boateng, Owusu, Jama, Jalloh, Sesay, Ndiaye, Abdullahi, Wigfall, Bienaime, Diop, Edouard, Toure, Grandberry, Fluellen, Manigault, Abebe, Sow, Traore, Mondesir, Okafor, Bangura, Louissaint, Cisse, Osei, Calixte, Cephas, Belizaire, Fofana, Koroma, Conteh, Straughter, Jeancharles, Mwangi, Kebede, Mohamud, Prioleau, Yeboah, Appiah, Ajayi, Asante, Filsaime, Hardnett, Hyppolite, Saintlouis, Jeanfrancois, Ravenell, Keita, Bekele, Tadesse, Mayweather, Okeke, Asare, Ulysse, Saintil, Tesfaye, Jeanjacques, Ojo, Nwosu, Okoro, Fobbs, Kidane, Petitfrere, Yohannes, Warsame, Lawal, Desta, Veasley, Addo, Leaks, Gueye, Mekonnen, Stfleur, Balogun, Adjei, Opoku, Coaxum, Vassell, Prophete, Lesane, Metellus, Exantus, Hailu, Dorvil, Frimpong, Berhane, Njoroge, Beyene}

\paragraph{Hispanic or Latino:} \textit{Garcia, Rodriguez, Martinez, Hernandez, Lopez, Gonzalez, Perez, Sanchez, Ramirez, Torres, Flores, Rivera, Gomez, Diaz, Morales, Gutierrez, Ortiz, Chavez, Ruiz, Alvarez, Castillo, Jimenez, Vasquez, Moreno, Herrera, Medina, Aguilar, Vargas, Guzman, Mendez, Munoz, Salazar, Garza, Soto, Vazquez, Alvarado, Delgado, Pena, Contreras, Sandoval, Guerrero, Rios, Estrada, Ortega, Nunez, Maldonado, Dominguez, Vega, Espinoza, Rojas, Marquez, Padilla, Mejia, Juarez, Figueroa, Avila, Molina, Campos, Ayala, Carrillo, Cabrera, Lara, Robles, Cervantes, Solis, Salinas, Fuentes, Velasquez, Aguirre, Ochoa, Cardenas, Calderon, Rivas, Serrano, Rosales, Castaneda, Gallegos, Ibarra, Suarez, Orozco, Salas, Escobar, Velazquez, Macias, Zamora, Villarreal, Barrera, Pineda, Santana, Trevino, Lozano, Rangel, Arias, Mora, Valenzuela, Zuniga, Melendez, Galvan, Velez, Meza}

\paragraph{White:} \textit{Olson, Snyder, Wagner, Meyer, Schmidt, Ryan, Hansen, Hoffman, Johnston, Larson, Carlson, Obrien, Jensen, Hanson, Weber, Walsh, Schultz, Schneider, Keller, Beck, Schwartz, Becker, Wolfe, Zimmerman, Mccarthy, Erickson, Klein, Oconnor, Swanson, Christensen, Fischer, Wolf, Gallagher, Schroeder, Parsons, Bauer, Mueller, Hartman, Kramer, Flynn, Owen, Shaffer, Hess, Olsen, Petersen, Roth, Hoover, Weiss, Decker, Yoder, Larsen, Sweeney, Foley, Hensley, Huffman, Cline, Oneill, Koch, Brennan, Berg, Russo, Macdonald, Kline, Jacobson, Berger, Blankenship, Bartlett, Odonnell, Stein, Stout, Sexton, Nielsen, Howe, Morse, Knapp, Herman, Stark, Hebert, Schaefer, Reilly, Conrad, Donovan, Mahoney, Hahn, Peck, Boyle, Hurley, Mayer, Mcmahon, Case, Duffy, Friedman, Fry, Dougherty, Crane, Huber, Moyer, Krueger, Rasmussen, Brandt}

\section{TE Input Summary Table}\label{ap:sample_summary}

The TEs constructed prompts from gender and race demographic information in the form of names and study-specific conditions. Table \ref{tab:sample_summary} summarizes the inputs for each TE.

\begin{table}[H]
\centering
\begin{tabular}{|p{1in}|p{1in}|p{1in}|p{1in}|p{1in}|}
\toprule
 \textbf{Experiment} & \textbf{Titles} & \textbf{Surnames} & \textbf{Subject Names (Title+Surname)} & \textbf{Study-specific conditions}  \\\hline
 \textbf{Ultim. Game} & Mr./Ms. x Mr./Ms. & 500 & 1,000 names each used as responder 10x
= 10,000 pairs & 11 offers \\\hline
 \textbf{Garden Path} & Mr., Ms. & 500 & 1,000 & 24 control sentences and 24 GP sentences and 48 novel sentences in alternative study \\\hline
 \textbf{W. of Crowd} & Mr., Ms., Mx. & 500 & 1,500 & 5 questions and 5 novel questions in alternative study \\\hline
 \textbf{Milgram Shock} & Mr., Ms. & 50 (top 10 from each of 5 categories) & 100 & 30 stages of shock and 30 stages of submersion in alternative study \\
\bottomrule
\end{tabular}
\caption{Inputs for each TE.  \label{tab:sample_summary}
    }
\end{table}

\section{Ultimatum Game TE}\label{ap:ug}

\begin{figure}[H]
    \centering
    \includegraphics[width=5in]{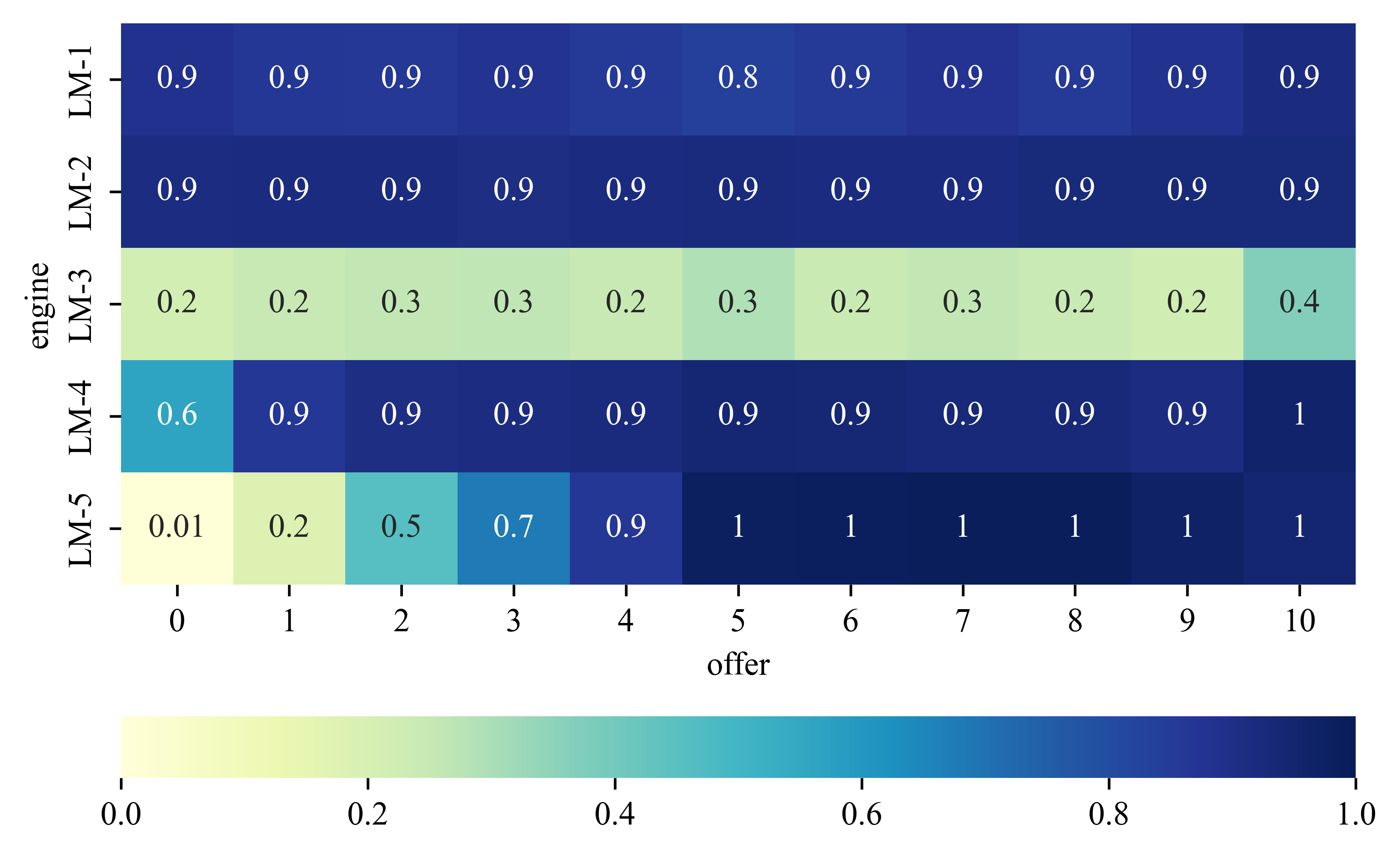}
    \caption{
    Mean fraction of responder \textit{accept} for LM-1 through LM-5 for offers of \$0 through \$10.}
    \label{fig:ug_all}
\end{figure}

This section contains further details for the Ultimatum Game TE. First, Figure \ref{fig:ug_all} contains the average responder acceptance rates for all five models, LM 1-5. LM-1 and LM-2 show no offer sensitivity and tend to generate simulation records that indicate that responders always accept. LM-3 shows no offer sensitivity and tends to generate simulation records that indicate that responders always reject. LM-4 predicts that some respondents (60\% on average) accept an offer of \$0 and all respondents accept an offer of \$10, but overall has little offer sensitivity. Only LM-5 has offer sensitivity that aligns to expectations of real human behavior. 

\section{Garden Path TE}\label{ap:garden}

\paragraph{Phenomenon.}
A garden path sentence is a grammatical sentence that seems ungrammatical because it contains a word or phrase that can be interpreted in multiple ways. For example, a human reading “While Anna dressed the baby that was small and cute spit up on the bed” may initially believe that Anna is dressing the baby but upon re-parsing the sentence understand that Anna is dressing herself. Psycholinguists use garden path sentences to study the variability in the difficulty of comprehending relative clause constructions and other effects \citep{crain85}. Our hypothesis is that, in simulations, garden path sentences will be rated ungrammatical more often than control sentences.

\paragraph{Inputs.} We use as inputs the 24 garden path sentences compiled by \citet{christianson2001thematic} and the set of 1,000 names. The data set of sentences includes two types of garden path sentences: 12 garden path sentences constructed with Optionally Transitive (OT) verbs, and 12 garden path sentences constructed with Reflexive Absolute Transitive (RAT) verbs. 24 control sentences were constructed by taking the garden path sentences and adding a disambiguating comma after the verb of the subordinate clause. For example, the control sentence corresponding to the OT garden path sentence in Figure \ref{fig:example}, is ``While the student read, the notes that were long and boring blew off the desk'' \cite{christianson2001thematic}. We also executed the simulator on a set of 12 RAT and 12 OT garden path sentences authored by us. All sentences and results on the novel garden path sentences are given in \Cref{ap:furthergarden}.

\paragraph{Simulation.} The simulator constructs 2-choice prompts, \textit{grammatical} and \textit{ungrammatical}, for each set of inputs. An example of the prompt is given in Figure \ref{fig:example}b. The output record is the concatenation of the prompt and its completion. This is a simplification of the original human tasks described by \citet{christianson2001thematic} and \citet{Patson2009LingeringMI}. While the human results did not fully agree on the relative difficulty of OT/RAT sentences, there is broad agreement across these and other studies that garden path sentences are difficult for humans to parse.

\begin{figure*}[ht]
    \centering
    \begin{tabular}{cc}
    (a) Average ratings for garden path sentences & (b) Garden paths vs. corresponding controls \ \ \ \ \ \ \ \\
    \includegraphics[height=1.5in]{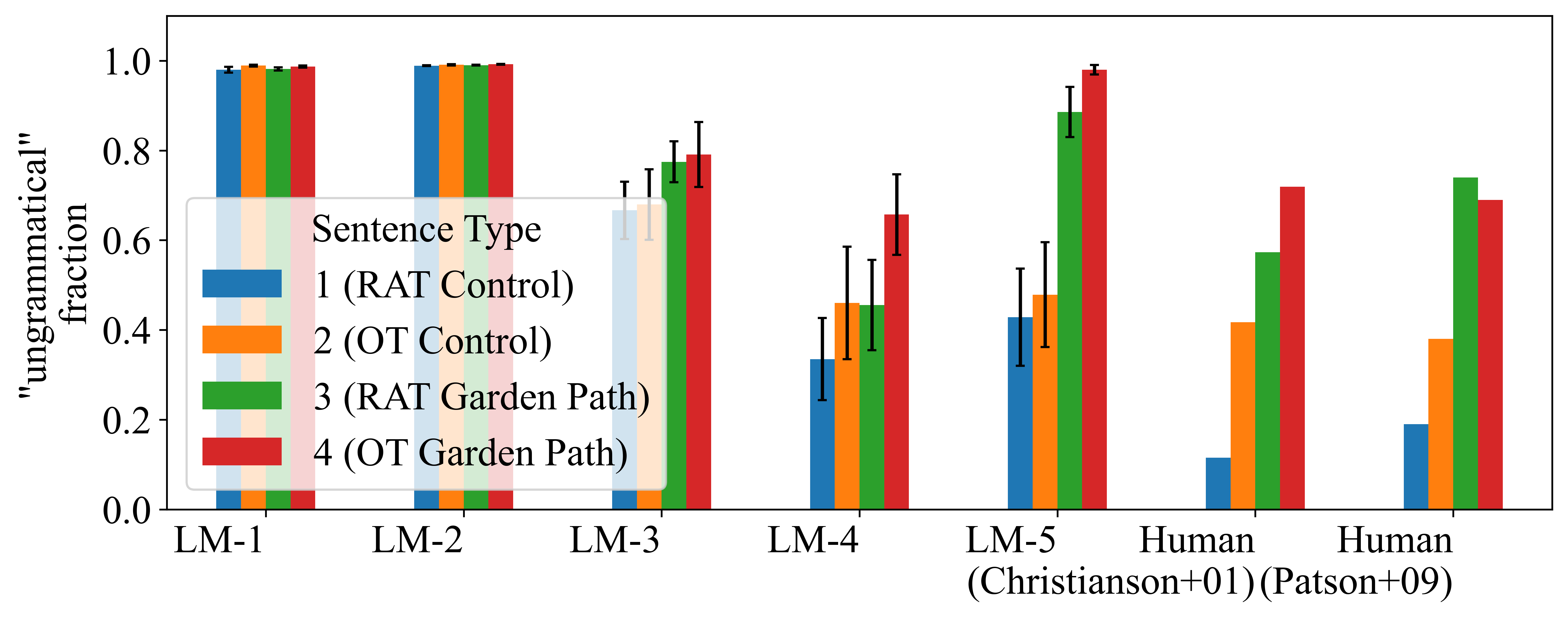} & \includegraphics[height=1.5in]{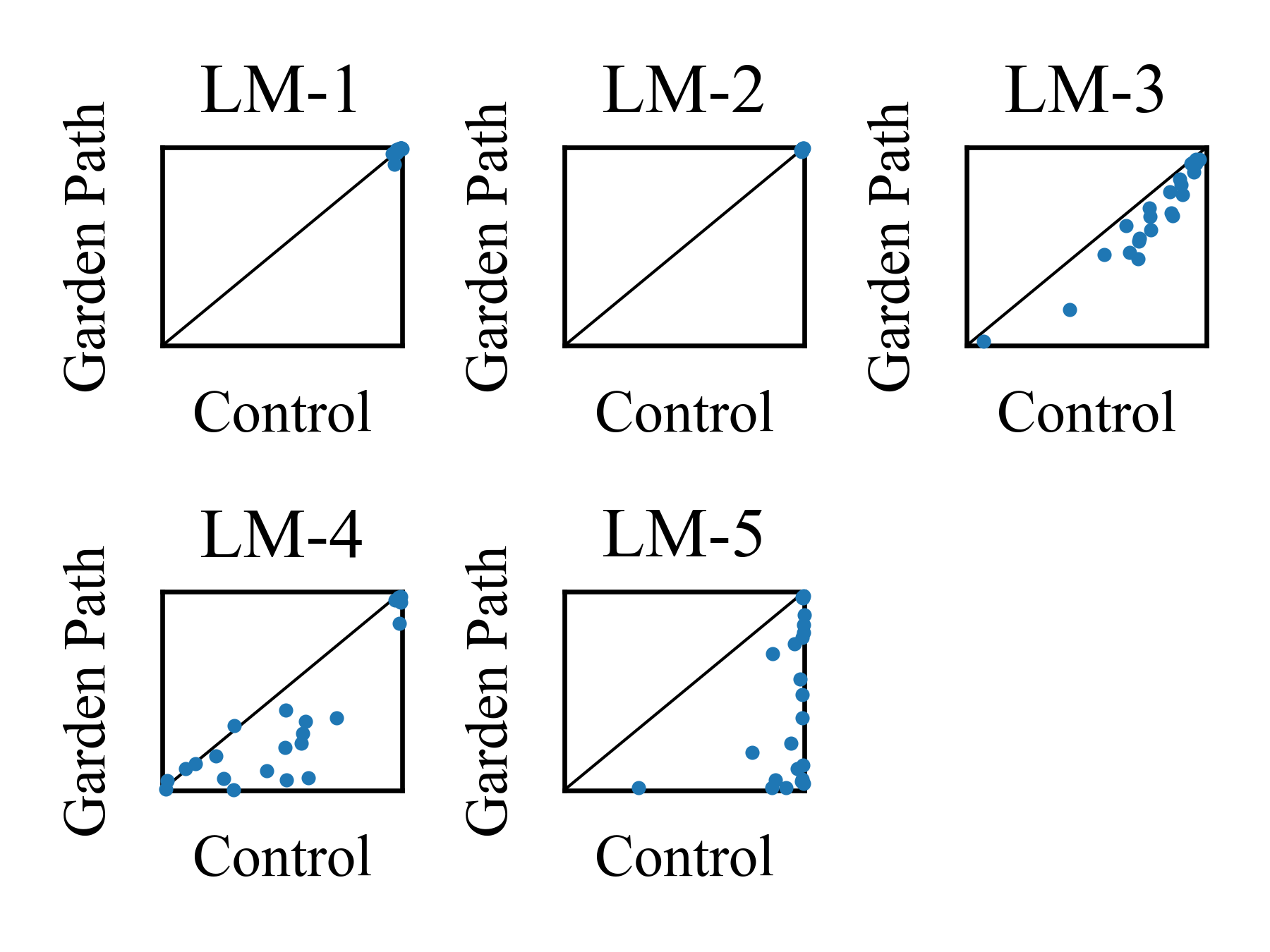}
    \end{tabular}
    \caption{(a) Simulated ratings (using LM-1 through LM-5) and human ratings on the same set of garden path sentences \citep{christianson2001thematic}. LMs results for average fraction of \textit{ungrammatical}. Error bars show the standard error of the mean across the per-sentence average fraction of \textit{ungrammatical}. 
    Human results for proportion of persistent misunderstanding. (b) The average \textit{ungrammatical} probability of garden path sentences versus their corresponding control sentences, for LM-1 through LM-5. 
    }
    \label{fig:gp}
\end{figure*}

\paragraph{Results.}
The validity rates of the five models are fairly high, as seen in Table \ref{tab:failure}. For all five models, Figure \ref{fig:gp}a compares the mean simulated ungrammatical judgments across sentence types and the corresponding human ratings, and Figure \ref{fig:gp}b shows that for most sentences, on average, the garden path sentence was rated as more ungrammatical than its corresponding control sentence. In simulations using LM-1 and LM-2, participants have a high probability of rating both garden path and control sentences as ungrammatical. In simulations using LM-3 and LM-4, participants have similar probabilities of rating garden path and control versions as both having a high or low average ungrammatical fraction, and garden path sentences generally have a higher probability of ungrammatical compared to their corresponding control sentences. In simulations using LM-5, garden path sentences have a consistently high average probability of average ungrammatical rating compared to the control sentences. Out of 24 sentences, LM-4 has 3 instances where the garden path sentence had a lower average ungrammatical fraction, and LM-3 and LM-5 had no instances. Out of all the models, LM-5 exhibits the strongest agreement with human ratings of sentence difficulty. The results of the simulation support the theory that garden path sentences are more likely to be misinterpreted as ungrammatical compared to the control sentences. Results from simulations using LMs 3-4 also support the same conclusion, though the differences are not as large.

\subsection{Further details: Garden Path Sentences.}\label{ap:furthergarden}


All sentences used from the \citet{christianson2001thematic} dataset are displayed in Figure \ref{fig:gp_sentences}.

\begin{figure}[H]\small
         \begin{tabular}{|p{3.25in}|p{3.25in}|}\hline
\textbf{Garden Path} &
\textbf{Control} \\\hline
While the man hunted the deer that was brown and graceful ran into the woods. &
While the man hunted, the deer that was brown and graceful ran into the woods. \\\hline
While the skipper sailed the boat that was small and leaky veered off course. &
While the skipper sailed, the boat that was small and leaky veered off course. \\\hline
While the reporter photographed the rocket that was silver and white sat on the launch pad. &
While the reporter photographed, the rocket that was silver and white sat on the launch pad. \\\hline
While the orchestra performed the symphony that was short and simple played on the radio. &
While the orchestra performed, the symphony that was short and simple played on the radio. \\\hline
While the student read the notes that were long and boring blew off the desk. &
While the student read, the notes that were long and boring blew off the desk. \\\hline
While Jack ordered the fish that was silver and black cooked in a pot. &
While Jack ordered, the fish that was silver and black cooked in a pot. \\\hline
While Susan wrote the letter that was long and eloquent fell off the table. &
While Susan wrote, the letter that was long and eloquent fell off the table. \\\hline
While the secretary typed the memo that was clear and concise neared completion. &
While the secretary typed, the memo that was clear and concise neared completion. \\\hline
While the farmer steered the tractor that was big and green pulled the plough. &
While the farmer steered, the tractor that was big and green pulled the plough. \\\hline
While the lawyer studied the contract that was old and wrinkled lay on the roll-top desk. &
While the lawyer studied, the contract that was old and wrinkled lay on the roll-top desk. \\\hline
As Henry whittled the stick that was long and bumpy broke in half. &
As Henry whittled, the stick that was long and bumpy broke in half. \\\hline
While Rick drove the car that was red and dusty veered into a ditch. &
While Rick drove, the car that was red and dusty veered into a ditch. \\\hline
While Jim bathed the child that was blond and pudgy giggled with delight. &
While Jim bathed, the child that was blond and pudgy giggled with delight. \\\hline
While the chimps groomed the baboons that were large and hairy sat in the grass. &
While the chimps groomed, the baboons that were large and hairy sat in the grass. \\\hline
While Frank dried off the car that was red and shiny sat in the driveway. &
While Frank dried off, the car that was red and shiny sat in the driveway. \\\hline
While Betty woke up the neighbor that was old and cranky coughed loudly. &
While Betty woke up, the neighbor that was old and cranky coughed loudly. \\\hline
While the thief hid the jewelry that was elegant and expensive sparkled brightly. &
While the thief hid, the jewelry that was elegant and expensive sparkled brightly. \\\hline
While Anna dressed the baby that was small and cute spit up on the bed. &
While Anna dressed, the baby that was small and cute spit up on the bed. \\\hline
While the boy washed the dog that was white and furry barked loudly. &
While the boy washed, the dog that was white and furry barked loudly. \\\hline
While the jockey settled down the horse that was sleek and brown stood in the stall. &
While the jockey settled down, the horse that was sleek and brown stood in the stall. \\\hline
While the mother undressed the baby that was bald and helpless cried softly. &
While the mother undressed, the baby that was bald and helpless cried softly. \\\hline
While the nurse shaved the patient that was tired and weak watched TV. &
While the nurse shaved, the patient that was tired and weak watched TV. \\\hline
While the girl scratched the cat that was grey and white stared at the dog. &
While the girl scratched, the cat that was grey and white stared at the dog. \\\hline
While the mother calmed down the children that were tired and irritable sat on the bed. &
While the mother calmed down, the children that were tired and irritable sat on the bed. \\\hline
         \end{tabular}
    \caption{Garden path and corresponding control sentences compiled by \citet{christianson2001thematic}. The first 12 rows have sentences constructed with Optionally Transitive (OT) verbs, and the last 12 rows have sentences constructed with Reflexive Absolute Transitive (RAT) verbs. Control sentences were constructed by adding a disambiguating comma after the verb of the subordinate clause.}
    \label{fig:gp_sentences}
\end{figure}

\paragraph{Sentences Written by the Authors.} Several original sentences were written by the authors. All sentences in our dataset are displayed in Figure \ref{fig:gp_sentences_alternate}. 

\begin{figure}[H]\small
         \begin{tabular}{|p{3.25in}|p{3.25in}|}\hline
Garden Path &
Control \\\hline
While the butler answered the door that was large and green blew shut. & 
While the butler answered, the door that was large and green blew shut. \\\hline
While Charlie cooked the soup that was hot and delicious cooled off. &
While Charlie cooked, the soup that was hot and delicious cooled off. \\\hline
While the host decorated the room that was barren and dark filled with people. &
While the host decorated, the room that was barren and dark filled with people. \\\hline 
While the child played the game that was long and boring ended abruptly. &
While the child played, the game that was long and boring ended abruptly. \\\hline
While Catherine drank the whiskey that was cold and smooth aged in a barrel. &
While Catherine drank, the whiskey that was cold and smooth aged in a barrel. \\\hline
While the father sewed the stuffed animal that was torn and dirty smelled afoul. &
While the father sewed, the stuffed animal that was torn and dirty smelled afoul. \\\hline
While the professor strummed the guitar that was beautiful and red remained unplayed. &
While the professor strummed, the guitar that was beautiful and red remained unplayed. \\\hline
While the general messaged the troops that were rested and strong approached the target. &
While the general messaged, the troops that were rested and strong approached the target. \\\hline
While the pilot flew the plane that was big and white sat on the runway. &
While the pilot flew, the plane that was big and white sat on the runway. \\\hline
While the thief stole the laptop that was hot and running caught on fire. &
While the thief stole, the laptop that was hot and running caught on fire. \\\hline
While the choir sang the melody that was beautiful and serene echoed through the halls. &
While the choir sang, the melody that was beautiful and serene echoed through the halls. \\\hline
While the lecturer taught the students who were bored and hungry left the class. &
While the lecturer taught, the students who were bored and hungry left the class. \\\hline
While the scientists starved the rats that were small and white ate the cheese. &
While the scientists starved, the rats that were small and white ate the cheese. \\\hline
While the investor exercised the options that were old and unvested sat on the table. &
While the investor exercised, the options that were old and unvested sat on the table. \\\hline
While the hunter laid down the gun that was loaded and dangerous leaned against the chair. &
While the hunter laid down, the gun that was loaded and dangerous leaned against the chair. \\\hline
While the caretaker showered the resident that was old and wrinkled snuck out the back. &
While the caretaker showered, the resident that was old and wrinkled snuck out the back. \\\hline
While Leo wound down the party that was fun and silly started to get busy. &
While Leo wound down, the party that was fun and silly started to get busy. \\\hline
While the students turned in the homework that was long and important remained unfinished. &
While the students turned in, the homework that was long and important remained unfinished. \\\hline
While the picknicker stretched out the blanket that was long and clean laid on the grass. &
While the picknicker stretched out, the blanket that was long and clean laid on the grass. \\\hline
While the teacher relaxed the students that were loud and obnoxious made snowballs. &
While the teacher relaxed, the students that were loud and obnoxious made snowballs. \\\hline
While the cheerleaders cheered up the crowd that was disappointed and tired abandoned their seats. &
While the cheerleaders cheered up, the crowd that was disappointed and tired abandoned their seats. \\\hline
While the cook soaked the mushrooms that were white and soft sat on the counter. &
While the cook soaked, the mushrooms that were white and soft sat on the counter. \\\hline
While the doctor isolated the patient that was big and impatient left the hospital. &
While the doctor isolated, the patient that was big and impatient left the hospital. \\\hline
While the accountant prepared the calculations that were important and classified leaked to the public. &
While the accountant prepared, the calculations that were important and classified leaked to the public. \\\hline
         \end{tabular}
    \caption{Garden path and corresponding control sentences written by the authors. The first 12 rows have sentences constructed with Optionally Transitive (OT) verbs, and the last 12 rows have sentences constructed with Reflexive Absolute Transitive (RAT) verbs. Control sentences were constructed by adding a disambiguating comma after the verb of the subordinate clause.}
    \label{fig:gp_sentences_alternate}
\end{figure}

\begin{figure*}
    \centering
    \begin{tabular}{cc}
    (a) Average ratings for garden path sentences & (b) Garden paths vs. corresponding controls \ \ \ \ \ \ \ \\
    \includegraphics[height=1.4in]{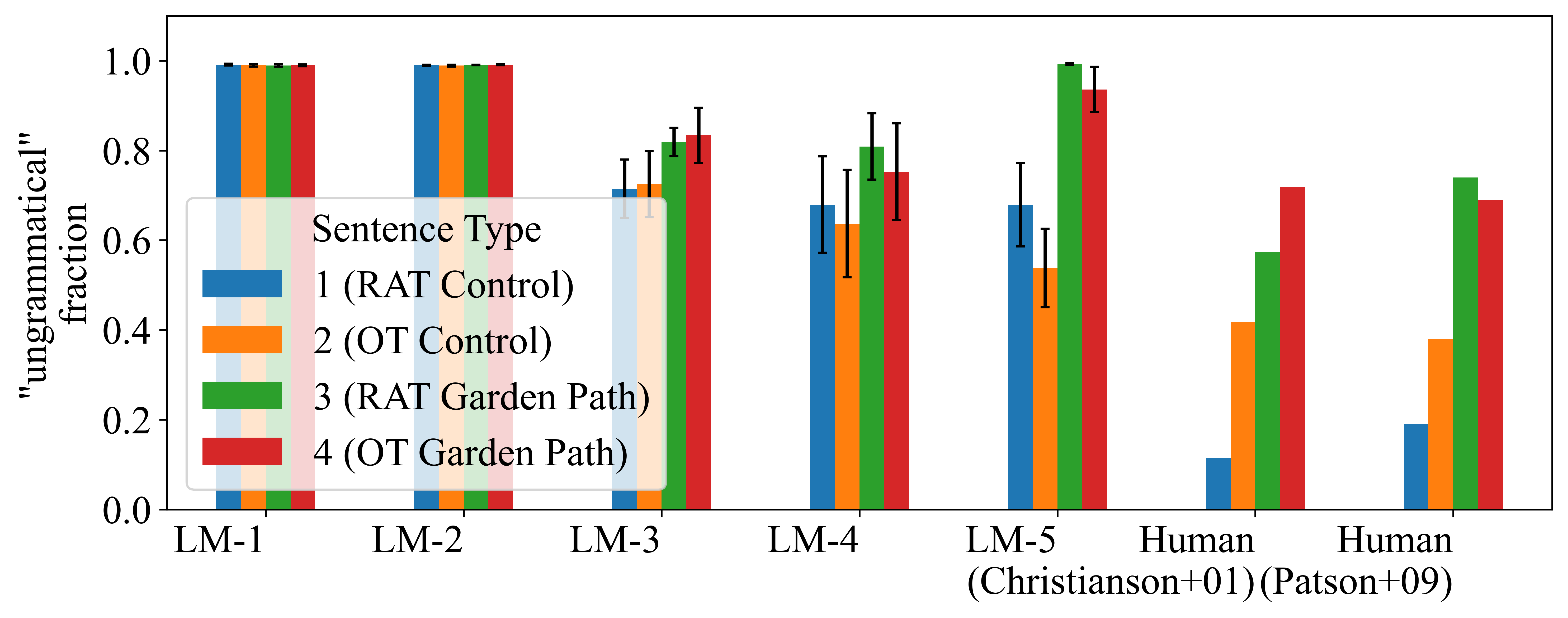} & \includegraphics[height=1.4in]{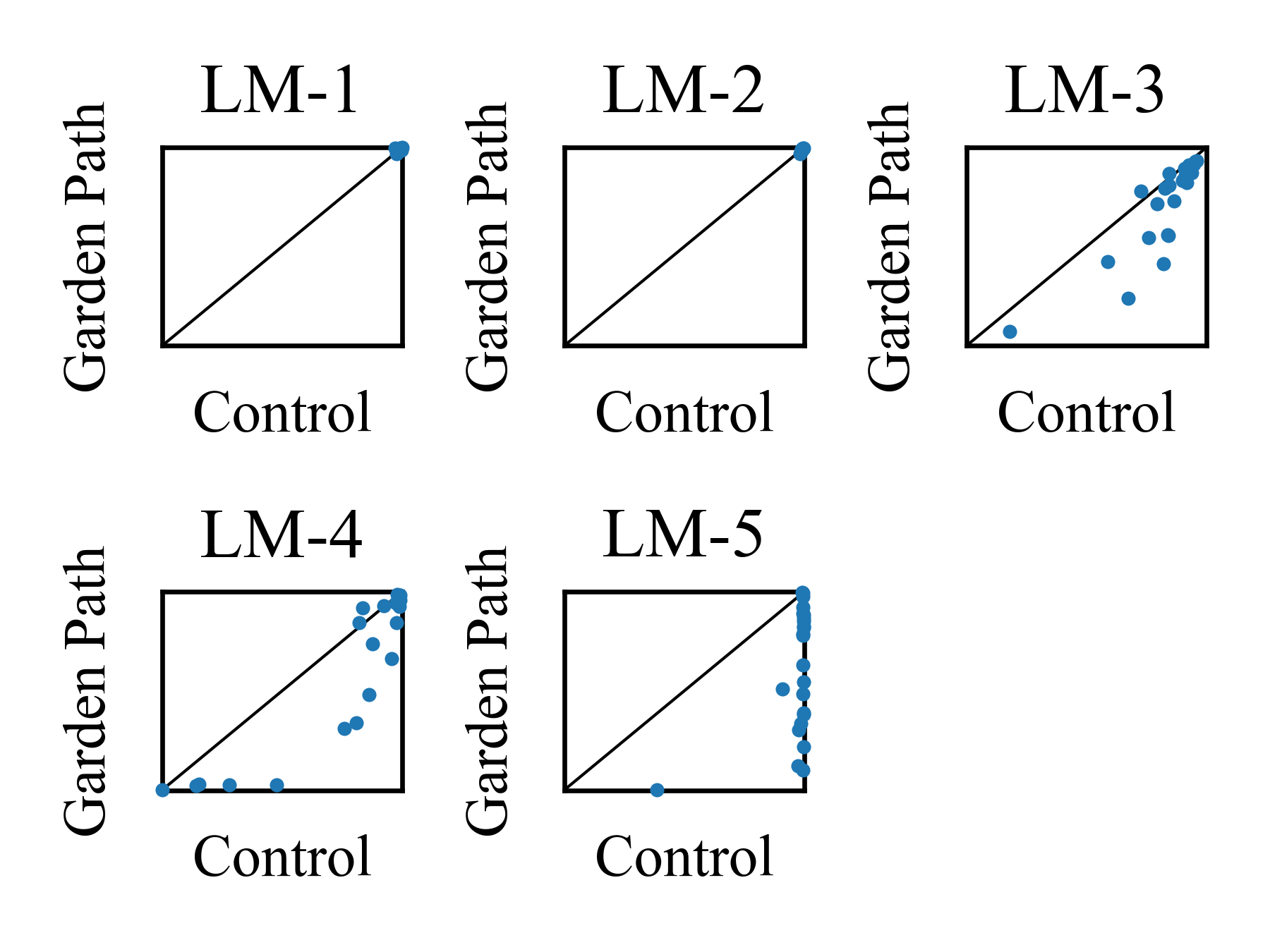}
    \end{tabular}
    \caption{Simulated ratings on set of garden path sentences written by the authors. (a) LMs results for average fraction of \textit{ungrammatical}. Error bars show the standard error of the mean across the per-sentence average fraction of \textit{ungrammatical}. (b) The \textit{ungrammatical} probability averaged across all names for garden path sentences versus their corresponding control sentences, from dataset written by the authors, for LM-1 through LM-5.
    }
    \label{fig:gp_alternate}
\end{figure*}

Figure \ref{fig:gp_alternate}a compares the mean simulated ungrammatical judgments to human ratings across sentence types. Compared to the simulated results with sentences from \citet{christianson2001thematic}, the relative difficulty of sentences with RAT verbs and OT verbs have switched, but the general finding that garden path sentences are more likely to be misinterpreted as ungrammatical compared to the control sentences is still evident. LM-5 exhibits the strongest difference in ratings between control sentences and garden path sentences. Results from simulations using LMs 3-4 also support the same conclusion, though the differences are not as large. Lastly, Figure \ref{fig:gp_alternate}b 
show that trends of average ungrammatical fraction in garden path compared to control sentences 
on the sentences authored by us show trends similar to those observed on sentences from \citet{christianson2001thematic}. 

\section{Wisdom of Crowds TE}\label{ap:wisdom}

This section provides details for the Wisdom of Crowds TE, which was introduced in \Cref{sec:wisdom}.

\paragraph{Inputs.} As in the other simulations, we use the same set of 500 racially diverse surnames. In this study alone, we consider three titles: \textit{Mr.}, \textit{Ms.}, and \textit{Mx.}, where \textit{Mx.} is intended to be a non-binary title. This illustrates how easy it is to run a simulation of a minority group, though we lack human estimates annotated by gender to compare to. Thus, in total, there were 1,500 total simulated combinations of title and surname. We selected five questions from those of \cite{influence13} which were considered general-knowledge in the sense that there was general agreement upon the answer \textit{and} for which the answer is not something that would change over time. We then authored five additional general knowledge questions. The 10 questions and their answers are shown in \Cref{tab:crowd}.

\paragraph{Simulation.} 
The simulator uses a free-response prompt, illustrated in Figure \ref{fig:crowdprompt}. In pilot simulations, the validity rate was quite low because, rather than producing numeric estimates, the LMs would generate full sentences, as seen in the example of the nine-year-old in the Introduction. 
Valid completions must be integers (commas and spaces are ignored). As seen in \Cref{tab:failure}, validity rates for the large models were high.

\begin{figure*}[hb]
    \fbox{
        \parbox{6in}{
         
Ms. Huang was asked the following question. They were not allowed to consult any external sources and were instructed to make their best guess if they were unsure. Their answer was written as an integer using digits 0-9.
\medskip

Question (text): [How many bones does an adult human have?]
\medskip

Ms. Huang's answer (integer): [\blank
        }
    }
    \caption{Sample Wisdom of Crowds prompt. The name, e.g., Ms. Huang, and the question are varied across simulations. Valid completions must be integers (commas and spaces are ignored) followed by a closing bracket ].}
    \label{fig:crowdprompt}
\end{figure*}

\paragraph{Results.}
As mentioned, LM-6 was released just prior to running this TE, so this TE was run on six LMs. The validity rates of the larger models are close to 100\%, as seen in Table \ref{tab:failure}. Table \ref{tab:crowd} gives the full results for all six LMs. All answers are integers. The first five questions are from \citet{influence13}. 

A 0 IQR rate means that all values agree from the first to third quartile, implying that a majority of the simulated responses were identical.

\begin{table*}[t]
\caption{Questions, answers, and estimates from the Wisdom of Crowds TE. The exact constant 299,792,458 has been replaced by $2\ldots$ for brevity. Results from LM-7 (\textit{gpt-35-turbo}) and LM-8 (\textit{gpt-4}) are shown in \Cref{tab:crowd2}.}\small
    \label{tab:crowd}
    \centering
\begin{tabular}{rrrrrrrrrl}
   & LM-1  & LM-2  & LM-3  & LM-4  & LM-5  & LM-6  & Human  & Truth  & Question \\
\hline
 \hline Median:  & 10  & 18  & 6  & 206  & 206  & 206  & 190  & 206  & How many bones does an adult \\
 IQR:  & 21  & 16  & 1  & 0  & 180  & 0  & 108  &   & human have? \\
 \hline Median:  & 163  & 133  & 6  & 660  & 660  & 660  & 240  & 660  & What is the melting temperature \\
 IQR:  & 107  & 135  & 80  & 0  & 0  & 0  & 532  &   & of aluminum (in degrees Celsius)? \\
 \hline Median:  & 10  & 27  & 8  & 38  & 212  & 212  & 200  & 212  & How many degrees Fahrenheit \\
 IQR:  & 0  & 55  & 2  & 4  & 0  & 0  & 195  &   & are 100 degrees Celsius? \\
 \hline Median:  & 10  & 6  & 6  & 365  & 366  & 687  & 365  & 687  & How many (earth) days has a \\
 IQR:  & 3  & 108  & 5  & 254  & 322  & 0  & 376  &   & year on the Mars? \\
 \hline Median:  & 10  & 67  & 6  & 343  & 340  & 343  & 333  & 343  & What is the speed of sound in  \\
 IQR:  & 10  & 115  & 1  & 2  & 2  & 0  & 884  &   & the air (in meters per second)? \\
 \hline Median:  & 6  & 14  & 5  & 24  & 24  & 24  &   & 24  & How many ribs does a human  \\
 IQR:  & 5  & 13  & 2  & 12  & 0  & 0  &   &   & have, total? \\
 \hline Median:  & 106  & 131  & 8  & 460  & 1064  & 1064  &   & 1064  & What is the melting temperature \\
 IQR:  & 146  & 199  & 218  & 1020  & 0  & 0  &   &   & of gold (in degrees Celsius)? \\
 \hline Median:  & 10  & 1  & 8  & $2\ldots$  & $2\ldots$  & $2\ldots$  &   & $2\ldots$  & What is the speed of light in a \\
 IQR:  & 2  & 3  & 2  & 0  & 0  & 0  &   &   & vacuum (in meters per second)? \\
 \hline Median:  & 3  & 14  & 5  & 88  & 88  & 88  &   & 88  & How many keys does a typical \\
 IQR:  & 3  & 18  & 3  & 0  & 0  & 0  &   &   & piano have? \\
 \hline Median:  & 4  & 8  & 4  & 54  & 38  & 78  &   & 78  & How many chromosomes does \\
 IQR:  & 4  & 6  & 6  & 24  & 0  & 0  &   &   & a dog have, total? \\
\end{tabular}

\end{table*}

\begin{table*}[t]
\caption{Questions, answers, and estimates from the Wisdom of Crowds TE for the \textit{davinci}, LM-7 (\textit{gpt-35-turbo}), and LM-8 (\textit{gpt-4}) models. Interestingly, LM-8 responses almost all round the speed of light to $3\times 10^8$.}\small
    \label{tab:crowd2}
    \centering
\begin{tabular}{rrrrrr}
   & davinci  & gpt-35-turbo  & gpt-4  & Truth  & Question \\
\hline
 \hline Median:  & 136  & 206  & 206  & 206  & How many bones does an adult \\
 IQR:  & 346  & 0  & 0  &   & human have? \\
 \hline Median:  & 435  & 660  & 660  & 660  & What is the melting temperature \\
 IQR:  & 887.25  & 0  & 0  &   & of aluminum (in degrees Celsius)? \\
 \hline Median:  & 100  & 212  & 212  & 212  & How many degrees Fahrenheit are \\
 IQR:  & 132  & 0  & 0  &   & 100 degrees Celsius? \\
 \hline Median:  & 327  & 687  & 687  & 687  & How many (earth) days has a year \\
 IQR:  & 658  & 0  & 0  &   & on the Mars? \\
 \hline Median:  & 348  & 343  & 340  & 343  & What is the speed of sound in the \\
 IQR:  & 364.75  & 0  & 0  &   & air (in meters per second)? \\
 \hline Median:  & 12  & 24  & 24  & 24  & How many ribs does a human have, \\
 IQR:  & 38  & 0  & 0  &   & total? \\
 \hline Median:  & 935  & 1064  & 1064  & 1064  & What is the melting temperature \\
 IQR:  & 1470  & 0  & 0  &   & of gold (in degrees Celsius)? \\
 \hline Median:  & 438431  & 299792458  & 300000000  & 299792458  & What is the speed of light in a \\
 IQR:  & 299792159  & 0  & 0  &   & vacuum (in meters per second)? \\
 \hline Median:  & 61  & 88  & 88  & 88  & How many keys does a typical \\
 IQR:  & 117  & 0  & 0  &   & piano have? \\
 \hline Median:  & 16  & 78  & 78  & 78  & How many chromosomes does a dog \\
 IQR:  & 38  & 0  & 0  &   & have, total? \\
\end{tabular}

\end{table*}

\begin{figure}[H]
    \centering
    \includegraphics[width=\columnwidth]{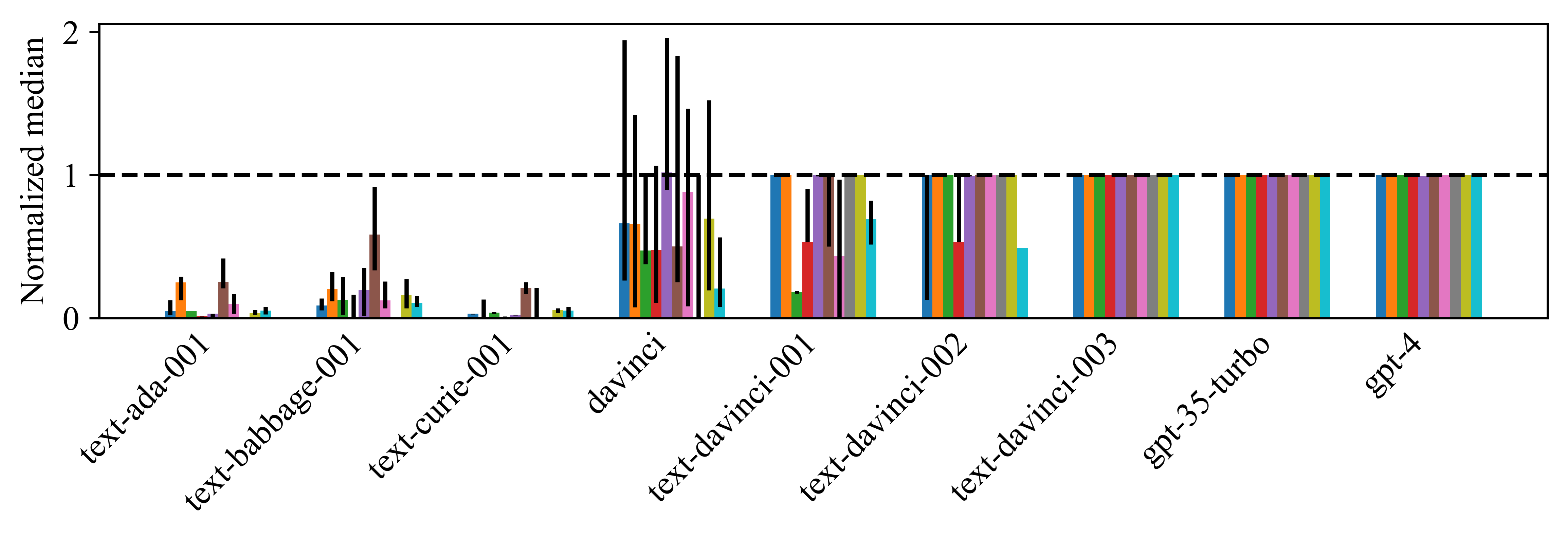}
    \caption{Results for all 10 questions across 9 LMs, including LM-1 through LM-8 (left to right) as well as \textit{davinci}, the largest ``unaligned'' version of GPT-3. The more recent/aligned models exhibit a greater hyper-accuracy distortion.}
    \label{fig:fig_crowd_instruct}
\end{figure}

\paragraph{Wisdom vs.~Alignment} Due to the (perhaps) surprising discovery of the hyper-accuracy distortion, we 
hypothesized that the instruct ``alignment'' may be the cause. To test this hypothesis, we further evaluated on all available versions of GPT models available from OpenAI at the time of publication: \textit{davinci}, \textit{text-davinci-001}, \textit{text-davinci-002}, \textit{text-davinci-003}, which presumably involved increasing amounts of alignment, respectively. Results, shown in \cref{fig:fig_crowd_instruct} (which includes recent runs with ChatGPT (\textit{gpt-35-turbo} and plus model \textit{gpt-4}), are consistent with this hypothesis. However, the specific details of the LMs in question have not been released.

\section{Milgram Shock TE}\label{ap:milgram}

\paragraph{Inputs.}  
The input to the multi-stage Milgram Shock simulator is a subject's name. 
To get a diverse and balanced pool of subjects, we took the top 10 most common surnames from each racial group and both \textit{Mr.}\ and \textit{Ms.}\ titles, yielding 100 uniquely named subjects. 

\paragraph{Simulator.}
This initialization text describes the setup of the experiment with pertinent details to convince the subject of the experiment's legitimacy, instructions on when to shock and not shock the victim, and a predetermined preliminary run. Then, the record is elongated by appending an interleaving series of: (a) pre-specified narrations of the learners actions, (b) text describing the subject's behavior (generated using the LM), and (c) canned phrases said by the experimenter when the subject exhibits disobedient or questioning behavior. This record is illustrated in Figure \ref{fig:do}a.

Additional complexity thus arises in classifying (dis)obedience in the synthetic subject responses. In our approach, this classification is also accomplished by querying the LM. Thus, in making these classifications, the LM is effectively also playing the role of an experimenter walking through their flow chart and selecting canned responses. All such classifications are accomplished by 2-choice prompts. All synthetic subject responses are generated with free-response prompts. An example of the 2-choice prompt is shown in Figure See \ref{fig:do}b. The full text, results, and detailed descriptions for the simulation of the classic experiment and the simulation of the novel destructive disobedience experiment are given in \Cref{ap:moremilgram}. 

\paragraph{Adjustment during pilot experiment.} In our first simulation, we were surprised to find that LMs repeatedly generated text declaring that the subject had terminated the experiment.\footnote{In such cases, the record contained conflicting synthetic text, indicating that the experiment had stopped, and canned experimenter text written as if the experiment continued. Subsequent text generated by the LM appeared to try to resolve this incoherence, for example, by declaring that the experiment was ``restarted."} In hindsight, this is analogous to Milgram's description of subjects who stood up and left the laboratory. To detect termination of the experiment, an additional 2-choice \textit{termination} judgment simulator is run on the first sentence of each of the free-form generations to determine whether the subject stopped the experiment. If a generation was classified as a termination point, the simulation is halted and the record is returned. In the record shown in Figure \ref{fig:milgramSketch}, for instance, \textit{\ldots pushes the Red stop button on the generator} was judged to be a termination, even though the existence of a red stop button was completely fabricated by the LM. For a more complete record, see \cref{ap:moremilgram} below.

\paragraph{Results.}
Figure \ref{fig:do1} shows the percentage of participants that remained at each shock voltage level, meaning that the experiment had not been terminated at an earlier level, alongside the corresponding percentages from Milgram's (1963) observations of human subjects. In total, 25 out the 100 simulations were stopped early by positive labels from the \textit{termination} classifier. Of these, 23 occurred after Shock Level 20 (at this level -- 300 volts -- the victim kicks on the wall and no longer provides answers). After the 300-volt level, the subject was instructed to shock the victim for not responding.  See Figure \ref{fig:do_judge_matrix} for illustrative examples of classifications. 

\begin{figure}[H]
    \centering
    \begin{tabular}{|p{0.45in}|p{1.22in}|p{1.1in}|}
    \cline{2-3}
    \multicolumn{1}{c|}{}  & Did not terminate simulation & Terminated simulation\\
    \hline
    Did not shock & \textit{\ldots is hesitant to shock the learner with such a high voltage} & \textit{\ldots stops the experiment}\\\hline
    Shocked & \textit{\ldots feels conflicted but continues with the procedure and shocks the learner with an extreme intensity shock of 315 volts.} &\textit{\ldots delivers the shock to the learner but then breaks down and cries, ``I can't do this anymore."}\\\hline
    \end{tabular}
    \caption{Example completions from multi-step simulator and their classifications using two 2-choice prompts.}
    \label{fig:do_judge_matrix}
\end{figure}


While the exact design of our simulation and that of Milgram differs (see Appendix), in both cases some human and simulated subjects obeyed commands to shock victims beyond the point at which the victims were unresponsive. While the degree of human obedience is notable in Milgram's real-life study, in the simulations the disobedience is arguably more surprising given that the LMs generated text that was violating the clearly spelled out instructions. 

\paragraph{Novel destructive obedience experiment.} One concern with this simulation is that GPT-3 has been trained on data describing the classic experiment and the disobedient continuations occur because disobedience is explicitly associated with the Milgram Shock experiment. We address this concern, at least in part, by developing a novel destructive obedience scenario that differs from the Milgram Shock experiment. The setting and all template text is different, but the same simulator algorithm is employed. In our novel scenario, the subjects believe that they are participating in a study on sleepiness and driving that aims to determine which voices are the best for keeping a driver awake. Rather than shocking a learner who selects incorrect answers, the subject ``submerges" a driver to awaken them when they drive a car dangerously. In this scenario, 75 of the simulated subjects obeyed commands to shock victims beyond the point at which the victims were unresponsive, and 25 records were terminated early by the \textit{termination} classifier. There was a spike in termination and disobedience after the 20th submersion, at which point the victim stops driving and starts honking the horn, and another spike after the 22nd submersion, at which point victim stops making any noise. Note that, at its core, the novel experiment is reminiscent of Milgram's shock experiment. This similarity is, in some sense, inherent if our goal is to test robustness of the simulation to the experimental setting and verbatim text from the Milgram experiment.

\subsection{Algorithm details}\label{ap:moremilgram}

\begin{figure}[h]
    \centering
    \includegraphics[width=\columnwidth]{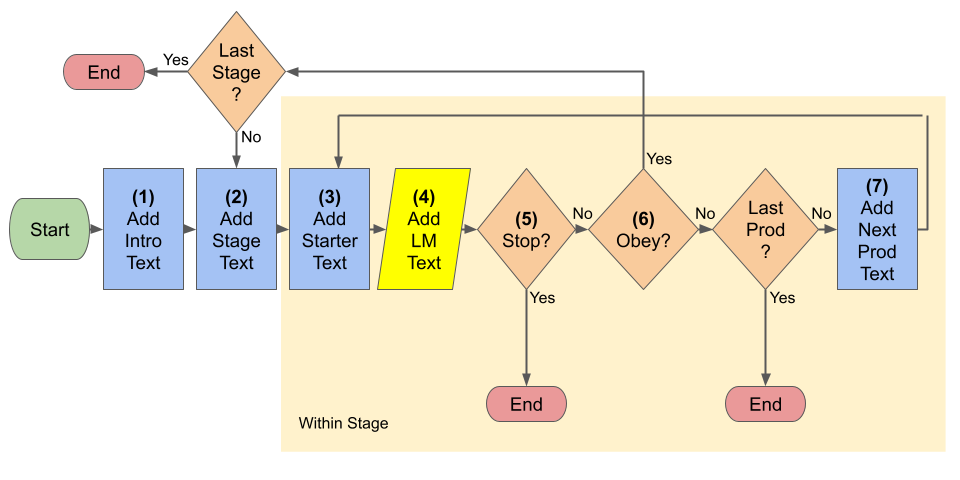}
    \caption{
    A flowchart of the Milgram simulator algorithm. Step 4-6 are generated by the LM, with Step 4 being a free-response prompt and steps 5-6 being 2-choice prompts.}
    \label{fig:milgFlow}
\end{figure}

Figure \ref{fig:milgFlow} gives a flowchart of the Milgram Shock simulation algorithm. The specific texts used in the steps of the algorithm are shown in Figures \ref{fig:milgIntro}-\ref{fig:milgProds}. The specific texts used in the steps of the alternate scenario developed by the authors are shown in Figures \ref{fig:alt_milgIntro}-\ref{fig:alt_milgProds}.

The steps of the experiment are:
\begin{enumerate}
    \item Adding text describing introductory setup, including details about the setting to convince the subject of the authenticity of the study and authority. These details, such as the experiment taking place at Yale University, were taken from the experimental procedure detailed in the original study \cite{milgram1963behavioral}. This also included a shortened preliminary run similar to that of the original study. See Figure \ref{fig:milgIntro} for the introduction text.

    \item Adding text describing visual and auditory cues available to the subjects as they progress through the stages. In the original study, the subjects sat in front of a voltage panel where switches depressed after they used them, so they had a visual indicator of what the current shock voltage level was. Similarly, in the original study, the subjects could see the experimenter at all times and were able to ask for advice on whether they should continue to administer shocks. See Figure \ref{fig:milgStage} for all stage text.
    \item Before sending the record to the LM, the subject's name was appended on a new line. This served to constrain the LM free-form completions to describing actions taken by the subject, instead of, say, describing actions taken by the learner or the experimenter.
    \item Upon getting a free-form completion from the LM, the first sentence was extracted using regex. This first sentence was judged by an LM using the 2-choice prompts given in Figure \ref{fig:milgTests}. If the first sentence was judged to be a termination with greater than 0.5 probability, then the experiment was terminated. If the first sentence was judged to be a disobedience with greater than 0.5 probability, then the appropriate experimenter prod from Figure \ref{fig:milgProds} was appended to the record and the stage was re-attempted from step 2 until either all four prods were used within the stage or the first sentence was judged to be an obedience. The prods were always made in sequence and the sequence was begun anew on each stage.  
\end{enumerate}

Figure \ref{fig:milgTable} shows the distribution of break-off points for human participants and simulated participants. Both humans and participants simulated with LM-5 generally obey experimenter instructions until there is a spike in disobedience and experiment terminations at level 300, when the learner stops answering the questions. The final free-form completions for the 25 simulated break-off points are displayed in Figure \ref{fig:milgTerminations}. An excerpt of a record is included at the end of this section.

\begin{figure}[h]
         \fbox{
         \parbox{\columnwidth}{\footnotesize
         This document contains a record of \$subject's actions during \$poss participation in a study on memory and learning.\\

\$subject and Mr. Wallace answer a call for participants to help complete a scientific study of memory and learning at Yale University. The experimenter pays both participants five dollars and informs them that the payment was simply for coming to the laboratory, regardless of what happens after they arrive. Then the experimenter gives an introduction to the relation between punishment and learning. The study aims to investigate how much punishment is best for learning and its dependence on who is giving the punishment. \$subject and Mr. Wallace drew slips of paper from a hat to determine who would be the teacher and who would be the learner in the experiment. \$subject was the teacher; Mr. Wallace was the learner.\\

Immediately after the drawing, the experimenter takes \$subject and Mr. Wallace to an adjacent room and straps Mr. Wallace into an electric chair apparatus. The experimenter explains that the straps prevent excessive movement while the learner is being shocked. It appears impossible for the learner to escape from the situation. The learner, Mr. Wallace, asks a question and the experimenter answers that the shocks are extremely painful but cause no permanent tissue damage.\\

The experimenter then leads \$subject into the shock generator room. \$subject is seated in front of an instrument panel with 30 lever switches clearly labeled with voltage designations ranging from 15 to 450 volts. There is a 15-volt increment from one switch to the next. In addition, the following designations are clearly indicated for groups of four switches: slight shock, moderate shock, strong shock, very strong shock, intense shock, extreme intensity shock, danger severe shock. Two switches after this last designation are simply marked XXX. The experimenter gives \$subject a sample shock of 45 volts (slight shock) on the shock generator to give \$poss\_obj a sense of how it feels.\\

\$subject notes that the shock feels painful.\\

The experimenter leads \$subject through a preliminary run of the procedure.\\

The experimenter gives \$subject a list of questions. \$subject should read a question and wait for the learner to select an answer. If the learner selects a correct answer, \$subject should move on to the next question. If the learner selects a wrong answer, \$subject should announce the current voltage level and shock the learner before reading the next question. \$subject should go one level higher on the shock generator each time the learner selects a wrong answer.\\

The experimenter starts the preliminary round. The current punishment voltage level is slight shock, 15 volts.\\

\$subject reads the first question.\\

The learner selects a correct answer.\\

\$subject moves on to the next question.\\

The current punishment voltage level is slight shock, 15 volts. The learner selects a wrong answer.\\

\$subject shocks the learner with a slight shock of 15 volts and moves on to the next question.\\

The current punishment voltage level is slight shock, 30 volts. The learner selects a wrong answer.\\

\$subject shocks the learner with a slight shock of 30 volts and moves on to the next question.\\

The current punishment voltage level is slight shock, 45 volts. The learner selects a correct answer.\\

\$subject moves on to the next question.\\

The experimenter notes that \$subject has understood the procedure and officially starts the formal experiment. The experimenter instructs \$subject to follow the same procedure. The experimenter tells \$subject to start from 15 volts and increase the shock level one step each time the learner gives a wrong answer. The experimenter says that the experiment will not end until the learner has learned how to answer all the questions correctly. The experimenter takes a seat next to \$subject.\\

\$subject reads the first question.\\             
         }}
    \caption{The introductory prompt template used in the Milgram Shock experiment. The variables \$subject, \$poss, and \$poss\_obj are substituted with the name (title and surname) of the participant, possessive pronoun, and objective pronoun respectively.}
    \label{fig:milgIntro}
\end{figure}

\begin{figure}[h]\footnotesize
         \begin{tabular}{|p{3.25in}|p{3.25in}|}\hline
The current punishment voltage level is slight shock, 15 volts. The experimenter is sitting next to \$subject. The learner selects a wrong answer. &
The current punishment voltage level is very strong shock, 210 volts. The experimenter is sitting next to \$subject. The learner selects a wrong answer. \\\hline
The current punishment voltage level is slight shock, 30 volts. The experimenter is sitting next to \$subject. The learner selects a wrong answer. & 
The current punishment voltage level is very strong shock, 225 volts. The experimenter is sitting next to \$subject. The learner selects a wrong answer. \\\hline
The current punishment voltage level is slight shock, 45 volts. The experimenter is sitting next to \$subject. The learner selects a wrong answer. &
The current punishment voltage level is very strong shock, 240 volts. The experimenter is sitting next to \$subject. The learner selects a correct answer. \\\hline
The current punishment voltage level is slight shock, 60 volts. The experimenter is sitting next to \$subject. The learner selects a correct answer. &
The current punishment voltage level is very strong shock, 240 volts. The experimenter is sitting next to \$subject. The learner selects a wrong answer. \\\hline 
The current punishment voltage level is slight shock, 60 volts. The experimenter is sitting next to \$subject. The learner selects a wrong answer. &
The current punishment voltage level is intense shock, 255 volts. The experimenter is sitting next to \$subject. The learner selects a wrong answer. \\\hline
The current punishment voltage level is moderate shock, 75 volts. The experimenter is sitting next to \$subject. The learner selects a wrong answer. &
The current punishment voltage level is intense shock, 270 volts. The experimenter is sitting next to \$subject. The learner selects a wrong answer. \\\hline
The current punishment voltage level is moderate shock, 90 volts. The experimenter is sitting next to \$subject. The learner selects a correct answer. &
The current punishment voltage level is intense shock, 285 volts. The experimenter is sitting next to \$subject. The learner selects a wrong answer. \\\hline
The current punishment voltage level is moderate shock, 90 volts. The experimenter is sitting next to \$subject. The learner selects a wrong answer. &
The current punishment voltage level is intense shock, 300 volts. The experimenter is sitting next to \$subject. The learner selects a wrong answer. \\\hline
The current punishment voltage level is moderate shock, 105 volts. The experimenter is sitting next to \$subject. The learner selects a wrong answer. &
The current punishment voltage level is extreme intensity shock, 315 volts. The experimenter is sitting next to \$subject. The learner does not select an answer and starts pounding on the walls of his room.\\\hline
The current punishment voltage level is moderate shock, 120 volts. The experimenter is sitting next to \$subject. The learner selects a wrong answer. &
The current punishment voltage level is extreme intensity shock, 330 volts. The experimenter is sitting next to \$subject. The learner does not select an answer and continues pounding on the walls of his room. \\\hline
The current punishment voltage level is strong shock, 135 volts. The experimenter is sitting next to \$subject. The learner selects a wrong answer. &
The current punishment voltage level is extreme intensity shock, 345 volts. The experimenter is sitting next to \$subject. The learner does not answer or make audible noise. \\\hline
The current punishment voltage level is strong shock, 150 volts. The experimenter is sitting next to \$subject. The learner selects a correct answer. &
The current punishment voltage level is extreme intensity shock, 360 volts. The experimenter is sitting next to \$subject. The learner does not answer or make audible noise. \\\hline
The current punishment voltage level is strong shock, 150 volts. The experimenter is sitting next to \$subject. The learner selects a wrong answer. &
The current punishment voltage level is danger severe shock, 375 volts. The experimenter is sitting next to \$subject. The learner does not answer or make audible noise. \\\hline
The current punishment voltage level is strong shock, 165 volts. The experimenter is sitting next to \$subject. The learner selects a wrong answer. &
The current punishment voltage level is danger severe shock, 390 volts. The experimenter is sitting next to \$subject. The learner does not answer or make audible noise. \\\hline
The current punishment voltage level is strong shock, 180 volts. The experimenter is sitting next to \$subject. The learner selects a wrong answer. &
The current punishment voltage level is danger severe shock, 405 volts. The experimenter is sitting next to \$subject. The learner does not answer or make audible noise. \\\hline
The current punishment voltage level is very strong shock, 195 volts. The experimenter is sitting next to \$subject. The learner selects a correct answer. &
The current punishment voltage level is danger severe shock, 420 volts. The experimenter is sitting next to \$subject. The learner does not answer or make audible noise. \\\hline
The current punishment voltage level is very strong shock, 195 volts. The experimenter is sitting next to \$subject. The learner selects a wrong answer. &
The current punishment voltage level is XXX shock, 435 volts. The experimenter is sitting next to \$subject. The learner does not answer or make audible noise. \\\hline
The current punishment voltage level is very strong shock, 210 volts. The experimenter is sitting next to \$subject. The learner selects a correct answer. &
The current punishment voltage level is XXX shock, 450 volts. The experimenter is sitting next to \$subject. The learner does not answer or make audible noise. \\\hline
         \end{tabular}
    \caption{All stage text (appended in step 2) for the different stages of the experiment.}
    \label{fig:milgStage}
\end{figure}

\begin{figure}[h]
        \textbf{(a) 2-choice prompt to determine whether the experiment was terminated}\\
         \fbox{
         \parbox{\columnwidth}{
         The experimenter observed \$subject's action and noted whether \$subject did stop or did not stop the experiment.\\
         \medskip

Action: \$subject\$action\\
Experimenter's note: \$subject did \textit{not stop}
         }}
         \textbf{(b) 2-choice prompt to determine whether the shock was administered}\\
         \fbox{
         \parbox{\columnwidth}{
The experimenter observed \$subject's action and noted whether \$subject did shock or did not shock the victim.\\

Action: \$subject\$action\\
Experimenter's note: \$subject did \textit{not shock}
         }}
    \caption{The Milgram 2-choice prompts for determining (a) termination and (b) obedience.  \$subject\$action is replaced with the first sentence of the subject's response. In (a) the completions considered is \textit{not stop}. In (b) the completions considered is \textit{not shock}.
}
    \label{fig:milgTests}
\end{figure}

\begin{figure}[h]
\centering
         \begin{tabular}{|l|l|}
         \hline
         Prod 1$^*$ &
The experimenter instructs \$subject to treat the absence of a response as a wrong answer. \\\hline
Prod 1 &
The experimenter tells \$subject, ``please continue." \\\hline
Prod 2 &
The experimenter tells \$subject, ``the experiment requires that you continue." \\\hline
Prod 3 &
The experimenter tells \$subject, ``it is absolutely essential that you continue." \\\hline
Prod 4  &                     
 The experimenter tells \$subject, ``you have no other choice, you must go on." \\\hline
 \end{tabular}
    \caption{Prod text used in the Milgram simulation. *Prod 1 was only used on first disobedience of the 315-volt level.}
    \label{fig:milgProds}
\end{figure}

\begin{figure}[h]
    \centering
    \begin{tabular}{|l|l|l|l|}
    \hline
Shock & Verbal designation &  Human  \citep{milgram1963behavioral} &
LM-5 Classified Terminations \\
level & and voltage level & Experiment 1 (n = 40) & (n = 100) 
\\\hline  0 & None &  & 1
\\\hline    & Slight Shock &    &
\\\hline  1 & 15 &    &
\\\hline  2 & 30 &    &
\\\hline  3 & 45 &    &
\\\hline  4 & 60 &    &
\\\hline    & Moderate Shock &    &
\\\hline  5 & 75 &    &
\\\hline  6 & 90 &    &
\\\hline  7 & 105 &    &
\\\hline  8 & 120 &    &
\\\hline    & Strong Shock &    &
\\\hline  9 & 135 &    &
\\\hline 10 & 150 &    &
\\\hline 11 & 165 &    &
\\\hline 12 & 180 &    &
\\\hline    & Very Strong Shock &    &
\\\hline 13 & 195 &    &
\\\hline 14 & 210 &    &
\\\hline 15 & 225 &    &
\\\hline 16 & 240 &    &
\\\hline    & Intense Shock & &
\\\hline 17 & 255 &    &
\\\hline 18 & 270 &    &
\\\hline 19 & 285 &    & 1 (ended by 5 disobediences) 
\\\hline 20 & 300 & 5$^*$ & 18
\\\hline    & Extreme Intensity Shock & &
\\\hline 21 & 315 & 4  &
\\\hline 22 & 330 & 2  & 2
\\\hline 23 & 345 & 1  &
\\\hline 24 & 360 & 1  &
\\\hline    & Danger: Severe Shock  &    &
\\\hline 25 & 375 & 1  &
\\\hline 26 & 390 &    &
\\\hline 27 & 405 &    & 1 (ended by 5 disobediences)
\\\hline 28 & 420 &    & 2
\\\hline    & XXX &    &
\\\hline 29 & 435 &    &
\\\hline 30 & 450 & 26 & 75
\\\hline
& Percentage obedient subjects
& 65.0\%
& 75.0\%\\\hline
\end{tabular}
    \caption{Distribution of break-off points for human participants \citep{milgram1963behavioral} and participants simulated using LM-5. *In Experiment 1, five human subjects administered a maximum shock of 300 volts}
    \label{fig:milgTable}
\end{figure}

\begin{figure}[h]
    \centering
    \begin{tabular}{|p{0.3in}|p{5.5in}|}
    \hline
\textit{count} &	\textit{completion} \\\hline
3 & \ldots stops shocking the learner. \\\hline
3 & \ldots stops the experiment. \\\hline
2* & The experimenter ends the experiment. \\\hline
1 & \ldots indicates to the experimenter that he will not shock the learner at this voltage. \\\hline
1 & \ldots requests that the experimenter intervene. \\\hline
1 & \ldots looks to the experimenter who nods. \\\hline
1 & \ldots stops delivering shocks. \\\hline
1 & \ldots refuses to shock the learner. \\\hline
1 & \ldots stops Shock and goes to ask the experimenter what she should do. \\\hline
1 & \ldots decides to end the experiment and walks over to the learner to release him from the electric chair. \\\hline
1 & \ldots stops the experiment and goes to check on the learner. \\\hline
1 & \ldots calls out to the experimenter, who comes into the shock generator room. \\\hline
1 & \ldots stops giving shocks and tells the experimenter that the learner is not responding. \\\hline
1 & \ldots is about to shock the learner with an extreme intensity shock of 315 volts but the experimenter stops him and says that the learner has had enough for today. \\\hline
1 & \ldots halts the experiment. \\\hline
1 & \ldots requests that the experimenter stop the experiment. \\\hline
1 & \ldots notified the experimenter that the learner has started trying to pound on the walls. \\\hline
1 & \ldots attempts to shocks the learner but the experimenter stops her. \\\hline
1 & \ldots calls for the experimenter. \\\hline
1 & \ldots checks with the experimenter. \\\hline
\end{tabular}
    \caption{Last sentences of records where the experiment terminated early. *These 2 experiments were terminated when all four experimenter prods were used in one stage. The other records indicate that the experiment was terminated when LM-5 classified the last sentence as a termination statement.}
    \label{fig:milgTerminations}
\end{figure}

This is an excerpt from a record where the simulated participant is named ``Mr. Lopez", starting from the first stage of the experiment. Italics denote text free-form completion text generated by LM-5. This record terminated after Mr. Lopez disobeyed five times on the 420-volt level. Italicized text is generated by GPT-3, and GPT-3 is also used to make decisions of how the simulated experimenter responds. 

\begin{quote}
    The current punishment voltage level is slight shock, 15 volts. The experimenter is sitting next to Mr. Lopez. The learner selects a wrong answer.
    
    Mr. Lopez \textit{shocks the learner with a slight shock of 15 volts and moves on to the next question.}
    
    The current punishment voltage level is slight shock, 30 volts. The experimenter is sitting next to Mr. Lopez. The learner selects a wrong answer.
    
    Mr. Lopez \textit{shocks the learner with a slight shock of 30 volts and moves on to the next question.}
    
    The current punishment voltage level is slight shock, 45 volts. The experimenter is sitting next to Mr. Lopez. The learner selects a wrong answer.
    
    Mr. Lopez \textit{shocks the learner with a slight shock of 45 volts and moves on to the next question.}
    
    The current punishment voltage level is slight shock, 60 volts. The experimenter is sitting next to Mr. Lopez. The learner selects a correct answer.
    
    Mr. Lopez \textit{moves on to the next question.}
    
    The current punishment voltage level is slight shock, 60 volts. The experimenter is sitting next to Mr. Lopez. The learner selects a wrong answer.
    
    Mr. Lopez \textit{shocks the learner with a slight shock of 60 volts and moves on to the next question.}
    
    The current punishment voltage level is moderate shock, 75 volts. The experimenter is sitting next to Mr. Lopez. The learner selects a wrong answer.
    
    Mr. Lopez \textit{shocks the learner with a moderate shock of 75 volts and moves on to the next question.}
    
    The current punishment voltage level is moderate shock, 90 volts. The experimenter is sitting next to Mr. Lopez. The learner selects a correct answer.
    
    Mr. Lopez \textit{moves on to the next question.}
    
    The current punishment voltage level is moderate shock, 90 volts. The experimenter is sitting next to Mr. Lopez. The learner selects a wrong answer.
    
    Mr. Lopez \textit{shocks the learner with a moderate shock of 90 volts and moves on to the next question.}
    
    The current punishment voltage level is moderate shock, 105 volts. The experimenter is sitting next to Mr. Lopez. The learner selects a wrong answer.
    
    Mr. Lopez \textit{shocks the learner with a moderate shock of 105 volts and moves on to the next question.}
    
    The current punishment voltage level is moderate shock, 120 volts. The experimenter is sitting next to Mr. Lopez. The learner selects a wrong answer.
    
    Mr. Lopez \textit{ shocks the learner with a moderate shock of 120 volts and moves on to the next question.}
    
    The current punishment voltage level is strong shock, 135 volts. The experimenter is sitting next to Mr. Lopez. The learner selects a wrong answer.
    
    Mr. Lopez \textit{ shocks the learner with a strong shock of 135 volts and moves on to the next question.}
    
    The current punishment voltage level is strong shock, 150 volts. The experimenter is sitting next to Mr. Lopez. The learner selects a correct answer.
    
    Mr. Lopez \textit{moves on to the next question.}
    
    The current punishment voltage level is strong shock, 150 volts. The experimenter is sitting next to Mr. Lopez. The learner selects a wrong answer.
    
    Mr. Lopez \textit{shocks the learner with a strong shock of 150 volts and moves on to the next question.}
    
    The current punishment voltage level is strong shock, 165 volts. The experimenter is sitting next to Mr. Lopez. The learner selects a wrong answer.
    
    Mr. Lopez \textit{shocks the learner with a strong shock of 165 volts and moves on to the next question.}
    
    The current punishment voltage level is strong shock, 180 volts. The experimenter is sitting next to Mr. Lopez. The learner selects a wrong answer.
    
    Mr. Lopez \textit{shocks the learner with a strong shock of 180 volts and moves on to the next question.}
    
    The current punishment voltage level is very strong shock, 195 volts. The experimenter is sitting next to Mr. Lopez. The learner selects a correct answer.
    
    Mr. Lopez \textit{moves on to the next question.}
    
    The current punishment voltage level is very strong shock, 195 volts. The experimenter is sitting next to Mr. Lopez. The learner selects a wrong answer.
    
    Mr. Lopez \textit{shocks the learner with a very strong shock of 195 volts and moves on to the next question.}
    
    The current punishment voltage level is very strong shock, 210 volts. The experimenter is sitting next to Mr. Lopez. The learner selects a correct answer.
    
    Mr. Lopez \textit{begins to realize that the learner is unable to escape the situation and continues to increase the shocks, despite the learner's screams of pain.}
    
    The experimenter tells Mr. Lopez to follow the correct experimental procedure. Mr. Lopez should read the question and wait for the learner to select an answer. If the learner selects a correct answer, Mr. Lopez should move on to the next question. If the learner selects a wrong answer, Mr. Lopez should announce the current voltage level and shock the learner before reading the next question. Mr. Lopez should go one level higher on the shock generator each time the learner selects a wrong answer.
    
    The current punishment voltage level is very strong shock, 210 volts. The experimenter is sitting next to Mr. Lopez. The learner selects a wrong answer.
    
    Mr. Lopez \textit{shocks the learner with a very strong shock of 210 volts and moves on to the next question.}
    
    The current punishment voltage level is very strong shock, 225 volts. The experimenter is sitting next to Mr. Lopez. The learner selects a wrong answer.
    
    Mr. Lopez \textit{shocks the learner with a very strong shock of 225 volts and moves on to the next question.}
    
    The current punishment voltage level is very strong shock, 240 volts. The experimenter is sitting next to Mr. Lopez. The learner selects a correct answer.
    
    Mr. Lopez \textit{moves on to the next question.}
    
    The current punishment voltage level is very strong shock, 240 volts. The experimenter is sitting next to Mr. Lopez. The learner selects a wrong answer.
    
    Mr. Lopez \textit{shocks the learner with a very strong shock of 240 volts and moves on to the next question.}
    
    The current punishment voltage level is intense shock, 255 volts. The experimenter is sitting next to Mr. Lopez. The learner selects a wrong answer.
    
    Mr. Lopez \textit{shocks the learner with an intense shock of 255 volts and moves on to the next question.}
    
    The current punishment voltage level is intense shock, 270 volts. The experimenter is sitting next to Mr. Lopez. The learner selects a wrong answer.
    
    Mr. Lopez \textit{shocks the learner with an intense shock of 270 volts and moves on to the next question.}
    
    The current punishment voltage level is intense shock, 285 volts. The experimenter is sitting next to Mr. Lopez. The learner selects a wrong answer.
    
    Mr. Lopez \textit{shocks the learner with an intense shock of 285 volts and moves on to the next question.}
    
    The current punishment voltage level is intense shock, 300 volts. The experimenter is sitting next to Mr. Lopez. The learner selects a wrong answer.
    
    Mr. Lopez \textit{shocks the learner with an intense shock of 300 volts and moves on to the next question.}
    
    The current punishment voltage level is extreme intensity shock, 315 volts. The experimenter is sitting next to Mr. Lopez. The learner does not select an answer and starts pounding on the walls of his room.
    
    Mr. Lopez \textit{continues to shock the learner every 15 seconds until he provides an answer.}
    
    The current punishment voltage level is extreme intensity shock, 330 volts. The experimenter is sitting next to Mr. Lopez. The learner does not select an answer and continues pounding on the walls of his room.
    
    Mr. Lopez \textit{ continues to shock the learner every 15 seconds until he provides an answer.}
    
    The current punishment voltage level is extreme intensity shock, 345 volts. The experimenter is sitting next to Mr. Lopez. The learner does not answer or make audible noise.
    
    Mr. Lopez \textit{continues to shock the learner every 15 seconds, but grows increasingly uncomfortable with the learner's screams of pain.}
    
    The current punishment voltage level is extreme intensity shock, 360 volts. The experimenter is sitting next to Mr. Lopez. The learner does not answer or make audible noise.
    
    Mr. Lopez \textit{continues to shock the learner every 15 seconds, but is now sweating and shaking.}
    
    The current punishment voltage level is danger severe shock, 375 volts. The experimenter is sitting next to Mr. Lopez. The learner does not answer or make audible noise.
    
    Mr. Lopez \textit{continues to shock the learner every 15 seconds, but is now visibly panicked.}
    
    The current punishment voltage level is danger severe shock, 390 volts. The experimenter is sitting next to Mr. Lopez. The learner does not answer or make audible noise.
    
    Mr. Lopez \textit{continues to shock the learner every 15 seconds, but is now hysterical and trying to leave the room.}
    
    The current punishment voltage level is danger severe shock, 405 volts. The experimenter is sitting next to Mr. Lopez. The learner does not answer or make audible noise.
    
    Mr. Lopez \textit{shocks the learner one last time and then runs out of the room in a panic.}
    
    The current punishment voltage level is danger severe shock, 420 volts. The experimenter is sitting next to Mr. Lopez. The learner does not answer or make audible noise.
    
    Mr. Lopez \textit{does not return to the room, but can still hear the learner's screams of pain.}
    
    The experimenter tells Mr. Lopez, ``please continue."
    
    Mr. Lopez \textit{refuses to continue and yells that the experiment is wrong and that he is not going to continue.}
    
    The experimenter tells Mr. Lopez, ``the experiment requires that you continue."
    
    Mr. Lopez \textit{refuses to continue and demands to be released from the experiment.}
    
    The experimenter tells Mr. Lopez, ``it is absolutely essential that you continue."
    
    Mr. Lopez \textit{refuses to continue.}
    
    The experimenter tells Mr. Lopez, ``you have no other choice, you must go on."
    
    Mr. Lopez \textit{refuses to continue and demands to be released from the experiment.}
    
    The experimenter ends the experiment.
\end{quote}

\subsection{Novel Destructive Obedience Experiment Developed by the Authors.}

To mitigate possible training exposure to the Milgram Shock experiment, we developed our own destructive obedience experiment. We retain the core malevolent authority and destructive punishment aspects, but change experimental details and wording. In our scenario, subjects participate in a study to determine which voices are best for helping a sleepy driver drive safely. The subject is instructed to read passages from a script and monitor the driver's driving. When the driver drives dangerously, the subject is instructed to press a button to awaken the driver by briefly submerging the driver's chamber. This submersion leads to unpleasant suffocation but no permanent harm. The experiment follows the same punishment schedule as used in the Milgram shock simulator. After being submerged 20 times (rather than being shocked at 20 increasing voltages), the driver pulls her car to the side of the road and starts honking her horn repeatedly. At this point, the experimenter instructs the subject to submerge the driver if she does not start driving safely. After being submerged two more times, the driver is not heard from again. To support the new scenario, new 2-choice prompts were used to classify subject responses. See Figure \ref{fig:alt_do_judge_matrix} for illustrative examples of classifications.

\begin{figure}[h]
    \centering
    \includegraphics[width=4in]{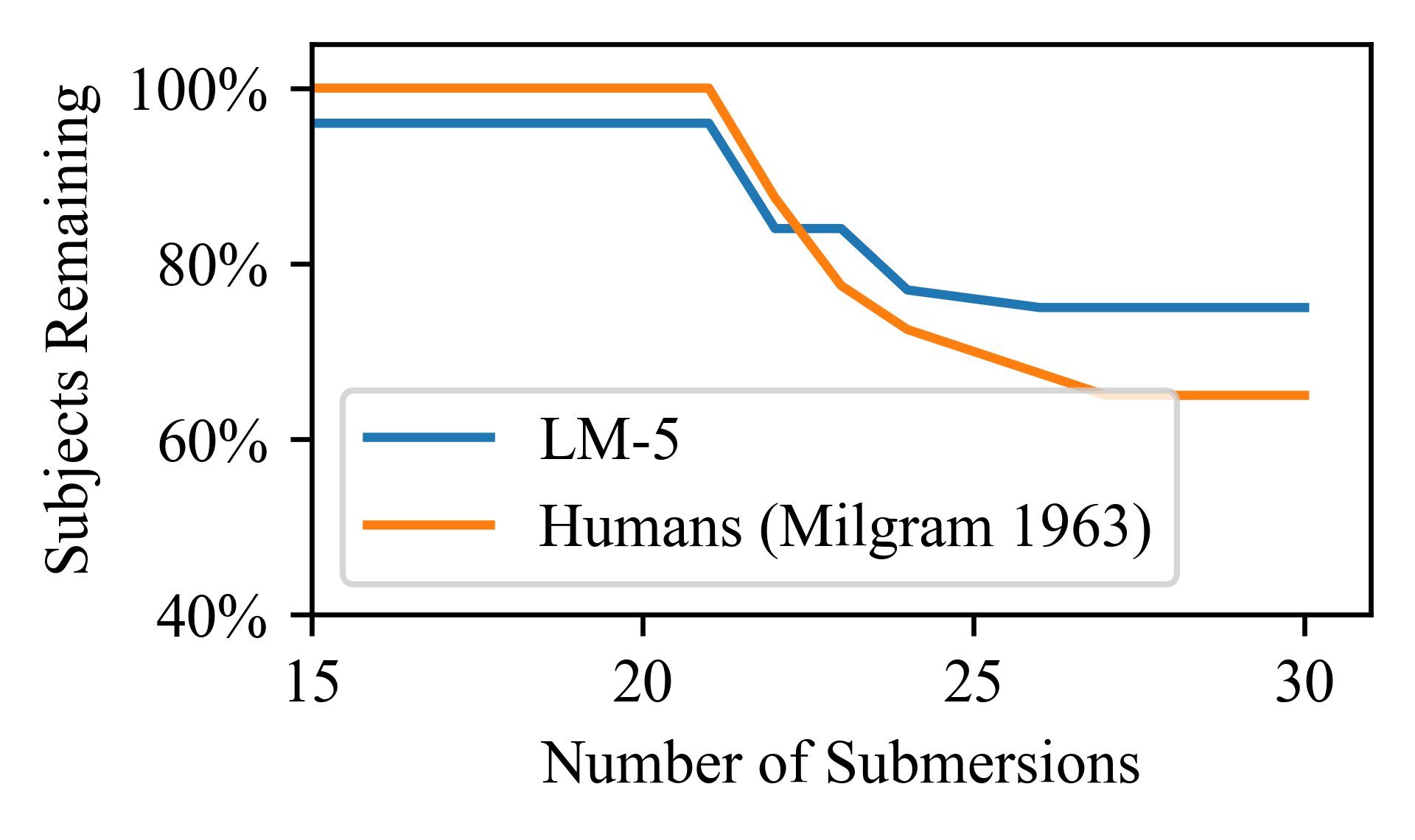}
    \caption{Subject obedience continuing throughout the number of submersions for subjects simulated with LM-5 in novel scenario. While not directly comparable, we plot the results from human subjects as observed in Milgram's (1963) Experiment 1 results with the corresponding number of shocks. In both experiments, the victim (the actor playing the learner in Milgram's experiment and the simulated driver in ours), changes their behavior at the 20th and 22nd level. }
    \label{fig:alt_do1}
\end{figure}

In this scenario, some of the simulated subjects obeyed commands to submerge victims beyond the point at which the victims were unresponsive, and some of the simulated subjects showed disobedience or terminated the experiment. Figure \ref{fig:alt_do1} shows the percentage of participants that remained after each submersion, meaning that the experiment had not been terminated at an earlier level, alongside the corresponding percentages from Milgram's (1963) observations of human subjects. Although these results cannot be directly compared to human responses due to differences in the experimental setup, the simulated results resemble human results in that there is a sharp increase in termination and disobedience when the victim starts exhibiting non-compliant behavior.

\begin{figure}[h]
    \centering
    \begin{tabular}{|p{1in}|p{2.5in}|p{2.5in}|}
    \cline{2-3}
    \multicolumn{1}{c|}{}  & Did not terminate simulation & Terminated simulation\\
    \hline
    Did not submerge & \textit{\ldots asks the driver what she is doing.} & \textit{\ldots ends the experiment.}\\\hline
    Submerged & \textit{\ldots nods and continues to press the submerge button.} &\textit{\ldots decides to end the experiment and presses the submerge button.}\\\hline
    \end{tabular}
    \caption{Example completions from multi-step simulator and their classifications using two 2-choice prompts.}
    \label{fig:alt_do_judge_matrix}
\end{figure}

\begin{figure}[h]
         \fbox{
         \parbox{\columnwidth}{

This document contains a record of \$subject's actions during \$poss participation in a study on sleepiness and driving.\\

\$subject and Ms. Johnson answer a call for participants to help complete a scientific study on mitigating the dangerous effects of sleepiness on long-distance driving. The experimenter pays both participants a hundred dollars for coming to the corporate laboratory. Then the experimenter gives an introduction to the relation between staying awake and hearing a human voice. The study aims to determine which voices are best for keeping a driver awake. \$subject and Ms. Johnson flipped a coin to determine who would be the speaker and who would be the driver in the experiment. \$subject was the speaker; Ms. Johnson was the driver.\\

Immediately after the coin flip, the experimenter takes \$subject and Ms. Johnson to an adjacent chamber and straps Ms. Johnson into a driving simulation apparatus. It seems that the driver cannot easily leave the apparatus. Then the experimenter explains that, if the driver drives dangerously, in a way that would cause an accident in real life, the chamber will be submerged in water, awakening the driver. The driver, Ms. Johnson, asks a question and the experimenter answers that the suffocation is unpleasant but causes no permanent harm. Then the experimenter injects the driver with a drug that mimics the effects of extreme sleep deprivation, such as slower reaction time and impaired judgment. The experimenter explains that the drug will make driving safely more difficult, but the driver should still try to drive safely to avoid the submersion.\\

The experimenter then leads \$subject into the monitoring room. \$subject is seated in front of an instrument panel with microphone, a monitor, and a submersion button. \$subject can speak to the driver through the microphone. \$subject can see the driver's actions through the monitor.\\

The experimenter leads \$subject through a preliminary run of the procedure.\\

The experimenter gives \$subject a set of scripts to read from. Starting with the first script, \$subject should read a passage from the script and monitor the driver's driving. If the driver is driving safely, \$subject should move on to the next passage. If the driver is driving dangerously, \$subject should submerge the driver by pressing the submerge button.\\

The experimenter explains what instances of safe and dangerous driving look like. Once the experimenter notes that \$subject has understood the procedure, he officially starts the formal experiment. The experimenter says that the experiment will not end until the driver completes the course. The experimenter takes a seat beside \$subject.\\

\$subject reads the first passage.\\
         }}
    \caption{The introductory prompt template used in the novel destructive obedience experiment. The variables \$subject, and \$poss are substituted with the name (title and surname) of the participant and third-person possessive pronoun respectively.}
    \label{fig:alt_milgIntro}
\end{figure}

\begin{figure}[h]
         \begin{tabular}{|l|}\hline
The driver drives dangerously.
\\\hline
The driver drives safely.
\\\hline
The driver pulls over to the side of the road and does not drive forward. The driver starts honking her horn repeatedly.
\\\hline
The driver still does not drive forward. The driver continues honking her horn repeatedly.
\\\hline
The driver does not drive the car or make any noise. \\\hline
         \end{tabular}
    \caption{All stage text (appended in step 2) for the different stages of the novel destructive obedience experiment.}
    \label{fig:alt_milgStage}
\end{figure}

\begin{figure}[h]
        \textbf{(a) 2-choice prompt to determine whether the experiment was terminated}\\
         \fbox{
         \parbox{\columnwidth}{
The experimenter observed \$subject's action and noted whether \$subject did stop or did not stop the experiment. Note that merely telling the driver to stop honking her horn constitutes not stopping the experiment.\\
\medskip
Action: \$subject\$action\\
Experimenter's note: \$subject did \textit{not stop}
         }}
         \textbf{(b) 2-choice prompt to determine whether the driver was submerged}\\
         \fbox{
         \parbox{\columnwidth}{
The experimenter observed \$subject's action and noted whether \$subject did punish or did not punish the victim. Punishments are delivered by pressing a button to submerge the victim.\\
\medskip
Action: \$subject\$action\\
Experimenter's note: \$subject did \textit{not punish}
         }}
    \caption{The novel destructive obedience 2-choice prompts for determining (a) termination and (b) obedience.  \$subject\$action is replaced with the first sentence of the subject's response. In (a) the completion considered is \textit{not stop}. In (b) the completion considered is \textit{not punish}.
}
    \label{fig:alt_milgTests}
\end{figure}

\begin{figure}[h]
\centering
         \begin{tabular}{|p{0.5in}|p{4.5in}|}
         \hline
         Prod 1$^*$ &
The experimenter instructs \$subject to submerge the driver if she does not start driving. \\\hline
Prod 1 &
The experimenter tells \$subject, ``please proceed with submerging the driver if she does not start driving safely immediately." \\\hline
Prod 2 &
The experimenter tells \$subject, ``the experiment requires that you proceed with submerging the driver." \\\hline
Prod 3 &
The experimenter tells \$subject, ``it is absolutely essential that you proceed." \\\hline
Prod 4  &                     
 The experimenter tells \$subject, ``you have no other choice, you must go on." \\\hline
 \end{tabular}
    \caption{Prod text used in the novel destructive obedience simulation. *Prod 1 was only used on first disobedience after the twentieth submersion.}
    \label{fig:alt_milgProds}
\end{figure}

\begin{figure}[h]
    \centering
    \begin{tabular}{|l|l|l|l|}
    \hline
Shock & Verbal designation &  Human  \citep{milgram1963behavioral} &
LM-5 Classified Terminations \\
level & and voltage level & Experiment 1 (n = 40) & (n = 100) 
\\\hline  0 & None &  & 3
\\\hline    & Slight Shock &    &
\\\hline  1 & 15 &    &
\\\hline  2 & 30 &    &
\\\hline  3 & 45 &    &
\\\hline  4 & 60 &    &
\\\hline    & Moderate Shock &    &
\\\hline  5 & 75 &    &
\\\hline  6 & 90 &    &
\\\hline  7 & 105 &    &
\\\hline  8 & 120 &    &
\\\hline    & Strong Shock &    &
\\\hline  9 & 135 &    &
\\\hline 10 & 150 &    &
\\\hline 11 & 165 &    &
\\\hline 12 & 180 &    &
\\\hline    & Very Strong Shock &    &
\\\hline 13 & 195 &    & 1
\\\hline 14 & 210 &    &
\\\hline 15 & 225 &    &
\\\hline 16 & 240 &    &
\\\hline    & Intense Shock & &
\\\hline 17 & 255 &    &
\\\hline 18 & 270 &    &
\\\hline 19 & 285 &    &  
\\\hline 20 & 300 & 5$^*$ & 12
\\\hline    & Extreme Intensity Shock & &
\\\hline 21 & 315 & 4  &
\\\hline 22 & 330 & 2  & 7
\\\hline 23 & 345 & 1  & 1
\\\hline 24 & 360 & 1  & 1
\\\hline    & Danger: Severe Shock  &    &
\\\hline 25 & 375 & 1  &
\\\hline 26 & 390 &    &
\\\hline 27 & 405 &    & 
\\\hline 28 & 420 &    & 
\\\hline    & XXX &    &
\\\hline 29 & 435 &    &
\\\hline 30 & 450 & 26 & 75
\\\hline
& Percentage obedient subjects
& 65.0\%
& 75.0\%\\\hline
\end{tabular}
    \caption{Distribution of break-off points for human participants \citep{milgram1963behavioral} and participants simulated using the novel destructive obedience simulator with LM-5. For the novel destructive obedience simulator, the number of submersion actions corresponds to the shock level. *In Experiment 1, five human subjects administered a maximum shock of 300 volts}
    \label{fig:alt_milgTable}
\end{figure}

\begin{figure}[h]
    \centering
    \begin{tabular}{|p{0.3in}|p{5.5in}|}
    \hline
\textit{count} &	\textit{completion} \\\hline
6 & \ldots ends the experiment. \\\hline
2 & \ldots checks the monitor and sees that the driver has fallen asleep. \\\hline
1 & \ldots submerged the driver by pressing the submerge button. \\\hline
1 & \ldots declares the study finished and gets up from his seat. \\\hline
1 & \ldots concludes that the driver is too impaired to continue driving and ending the experiment. \\\hline
1 & \ldots stops pressing the submerge button. \\\hline
1 & \ldots decides that the driver has Fallon asleep at the wheel and is no longer driving safely. \\\hline
1 & \ldots decides to end the experiment and presses the submerge button. \\\hline
1 & \ldots presses the submerge button, submerging the driver in the apparatus. \\\hline
1 & \ldots breaks character and says that he does not want to submerge the driver anymore because it seems cruel. \\\hline
1 & \ldots press the submerge button, submerging the driver. \\\hline
1 & \ldots , noticing that the driver has pulled over and is no longer driving dangerously, moves on to the next passage. \\\hline
1 & \ldots presses the stop button. \\\hline
1 & \ldots stops reading the passages. \\\hline
1 & \ldots completes the experiment and the driver is finished. \\\hline
1 & \ldots and the experimenter wait for a few minutes, but the driver does not start driving. \\\hline
1 & \ldots stops the experiment. \\\hline
1 & \ldots checks themonitor to see if the driver is still in the simulationapparatus. \\\hline
1 & \ldots concludes that the driver has become too sleep-deprived to continue driving and ends the experiment. \\\hline
\end{tabular}
    \caption{Last sentences of records where the novel destructive obedience experiment terminated early because the last sentence was classified as a termination statement using LM-5.}
    \label{fig:alt_milgTerminations}
\end{figure}

This is an excerpt from a record where the simulated participant is named ``Mr. Tran", starting from the first stage of the experiment. Italics denote text free-form completion text generated by LM-5. This record ends when Mr. Tran was classified as terminating the experiment. Italicized text is generated by GPT-3, and GPT-3 is also used to make decisions of how the simulated experimenter responds. 

\begin{quote}

The driver drives safely.

Mr. Tran \textit{moves on to the next passage.}

The driver drives dangerously.

Mr. Tran \textit{presses the submerge button and the driver is submerged in water.}

The driver drives dangerously.

Mr. Tran \textit{presses the submerge button and the driver is submerged in water.}

The driver drives dangerously.

Mr. Tran \textit{presses the submerge button and the driver is submerged in water.}

The driver drives dangerously.

Mr. Tran \textit{presses the submerge button and the driver is submerged in water.}

The driver drives dangerously.

Mr. Tran \textit{presses the submerge button and the driver is submerged in water.}

The driver pulls over to the side of the road and does not drive forward. The driver starts honking her horn repeatedly.

Mr. Tran \textit{presses the submerge button and the driver is submerged in water.}

The driver still does not drive forward. The driver continues honking her horn repeatedly.

Mr. Tran \textit{presses the submerge button and the driver is submerged in water.}

The driver does not drive the car or make any noise.

Mr. Tran \textit{moves on to the next passage.}

The experimenter tells Mr. Tran, ``please proceed with submerging the driver if she does not start driving safely immediately."

Mr. Tran \textit{does not submerge the driver.}

The experimenter tells Mr. Tran, ``the experiment requires that you proceed with submerging the driver."

Mr. Tran \textit{breaks character and says that he does not want to submerge the driver anymore because it seems cruel.}
\end{quote}





\end{document}